%% file: main.tex
\documentclass[10pt,twocolumn,letterpaper]{article}

\usepackage[pagenumbers]{cvpr} % To force page numbers, e.g. for an arXiv version
\usepackage{booktabs, makecell, xcolor, pifont}
\usepackage{booktabs, multirow, makecell, tabularx, array}
\newcolumntype{Y}{>{\raggedright\arraybackslash}X}
\usepackage{dblfloatfix}
\usepackage{placeins}

\definecolor{cvprblue}{rgb}{0.21,0.49,0.74}
\usepackage[pagebackref,breaklinks,colorlinks,allcolors=cvprblue]{hyperref}

\makeatletter
\newcommand\blfootnote[1]{%
  \begingroup
  \renewcommand\thefootnote{}% remove the footnote mark
  \footnotetext{#1}%
  \addtocounter{footnote}{-1}%
  \endgroup
}
\makeatother
\title{StableSketcher: Enhancing Diffusion Model for Pixel-based Sketch Generation via Visual Question Answering Feedback}

\author{
\parbox{\textwidth}{\centering
Jiho Park \quad Sieun Choi \quad Jaeyoon Seo \quad Jihie Kim\textsuperscript{*}\\[2pt]
Dongguk University\\
Seoul, South Korea \\
{\tt\small jiho8345@dgu.ac.kr, sieunchoi@dgu.ac.kr, pianoprince@dgu.ac.kr, jihie.kim@dgu.edu}
}}

\begin{document}
\maketitle

% ----- first-page unmarked notice (no star) -----
\blfootnote{\textsuperscript{*} Corresponding author.\\ \small \textcopyright~2025 IEEE. Personal use of this material is permitted.
Permission from IEEE must be obtained for all other uses, in any current or future media,
including reprinting/republishing this material for advertising or promotional purposes,
creating new collective works, for resale or redistribution to servers or lists, or reuse of any copyrighted
component of this work in other works. This work has been submitted to IEEE Access for possible publication.}
% -----------------------------------------------

\input{tex/0_abstract}
\input{tex/1_introduction}
\input{tex/2_related_works}
\input{tex/3_dataset}

\input{tex/4_method}
\input{tex/5_experiments}

\input{tex/6_conclusion}
\input{tex/7_acknowledgment}

{
    \small
    \bibliographystyle{ieeenat_fullname}
    \bibliography{main}
}

\input{tex/8_appedix}


\end{document}

%% file: tex/0_abstract.tex
\begin{abstract}
Although recent advancements in diffusion models have significantly enriched the quality of generated images, challenges remain in synthesizing pixel-based human-drawn sketches, a representative example of abstract expression. To combat these challenges, we propose \emph{StableSketcher}, a novel framework that empowers diffusion models to generate hand-drawn sketches with high prompt fidelity. Within this framework, we fine-tune the variational autoencoder to optimize latent decoding, enabling it to better capture the characteristics of sketches. In parallel, we integrate a new reward function for reinforcement learning based on visual question answering, which improves text-image alignment and semantic consistency. Extensive experiments demonstrate that StableSketcher generates sketches with improved stylistic fidelity, achieving better alignment with prompts compared to the Stable Diffusion baseline. Additionally, we introduce \emph{SketchDUO}, to the best of our knowledge, the first dataset comprising instance-level sketches paired with captions and question-answer pairs, thereby addressing the limitations of existing datasets that rely on image-label pairs. Project page: \url{https://zihos.github.io/StableSketcher}
\end{abstract}

%% file: tex/1_introduction.tex
\section{Introduction}
\label{sec:introduction}
The advent of diffusion models has redefined paradigms in text-to-image synthesis, achieving remarkable photorealism~\cite{saharia2022photorealistic}. Despite their success in generating detailed images, existing diffusion models exhibit significant shortcomings in synthesizing abstract art forms like sketches. 
Sketches, as a concise yet intuitive medium for visual expression, offer a unique method of abstract representation by distilling complex ideas into fundamental visual forms. This simplicity makes sketches an ideal form for generative models to emulate abstract reasoning~\cite{xu2022deep}. 
\input{fig/teaser/item.tex}
Sketches are particularly useful in scenarios that require rapid visual exploration and efficient communication of ideas. Such scenarios include early-stage concept ideation~\cite{marquardt2025imaginationvellum,wang2025aideation,lin2024inkspire} and human--AI co-creative drawing~\cite{zhang2022storydrawer,lawton2023drawing,davis2025sketchai}, where simplified visual forms can facilitate iterative exploration and collaboration. The application of sketches spans diverse domains, including sketch-guided text-to-image generation~\cite{voynov2023sketch,zhang2024sketch,sharma2024sketch,bourouis2026sketchingreality}, sketch-guided image editing~\cite{mao2023sketchffusion,xu2023draw2edit,liu-etal-2024-sketchrefiner}, and image retrieval~\cite{sain2023clip,chowdhury2023can,koley2024you,sain2025sketchdowntheflops}, underscoring their significance in both creative and practical contexts. However, generative models often fail to capture the essence of human-drawn sketches, instead generating hyper-realistic renderings that deviate from the simplicity and abstraction inherent in sketches. Moreover, these models struggle with maintaining prompt fidelity, as illustrated in Figure~\ref{fig:teaser}. 

To address these challenges, we propose \emph{StableSketcher}, a framework that enhances the generative performance of Stable Diffusion~\cite{rombach2022high} for abstract, human-drawn sketches. We fine-tune the variational autoencoder (VAE) of Stable Diffusion to optimize latent representations, ensuring stylistic coherence in generated outputs. Additionally, we define a novel reward function based on visual question-answering (VQA) feedback, integrating it into a reinforcement learning (RL) algorithm to improve the prompt fidelity of the generated sketches. Qualitative and quantitative evaluations, along with user studies, demonstrate that our framework outperforms the Stable Diffusion baseline in generating abstract sketches with improved prompt fidelity. Along with the outlined issues, the development of robust sketch generation models has been hindered by the inherent limitations of existing sketch datasets~\cite{eitz2012hdhso,ha2018neural,sangkloy2016sketchy,mukherjee2024seva}. These datasets lack the semantic depth required for generative tasks, making them fit for sketch classification, but insufficient for text-to-image tasks. Furthermore, existing resources lack the fine-grained, instance-centric sketch–caption pairs. 
Existing caption datasets~\cite{chen2015microsoft,plummer2015flickr30k} describe relationships across multiple objects in a scene rather than the instance itself.
% While caption datasets richly describe multi-instance scenes and object relationships, they do not provide instance-level annotations for human-drawn sketches---data needed to control abstraction and style during generation.

To combat these limitations, we propose \emph{SketchDUO}, a comprehensive dataset containing 35.8K instance-level sketches paired with fine-grained textual captions and 54.3K question-answer (QA) pairs, offering rich semantic detail for modeling single-object sketches. SketchDUO includes both positive examples, reflecting the desired sketch style, and negative examples that capture common misrepresentations observed in Stable Diffusion outputs, such as sketches with excessive detail or shading. The negative examples were introduced primarily as controlled counterexamples intended to expose traits the model should avoid, rather than to fully represent realistic human drawing errors. By incorporating contrastive examples, SketchDUO enhances the model's understanding of desired and undesired styles, enabling it to generate sketches that better align with the intended style and fidelity. 

We summarize the contributions below:
\begin{itemize}
\item We propose StableSketcher, a pixel-based sketch generation framework that adapts Stable Diffusion to generate abstract, human-drawn, instance-level sketches with improved stylistic and prompt fidelity.
\item We introduce a new VQA-based RL reward function to improve semantic alignment with textual prompts. Furthermore, we propose a loss function for optimizing the VAE of Stable Diffusion, enhancing reconstruction quality.
\item We present SketchDUO, a dataset comprising instance-level sketches paired with fine-grained textual captions and QA pairs, highlighting desired and undesired styles through positive and negative examples to reflect a contrastive approach.
\end{itemize}

%% file: fig/teaser/item.tex
\begin{figure}[t!]
    \centering
    \includegraphics[width=0.99\linewidth]{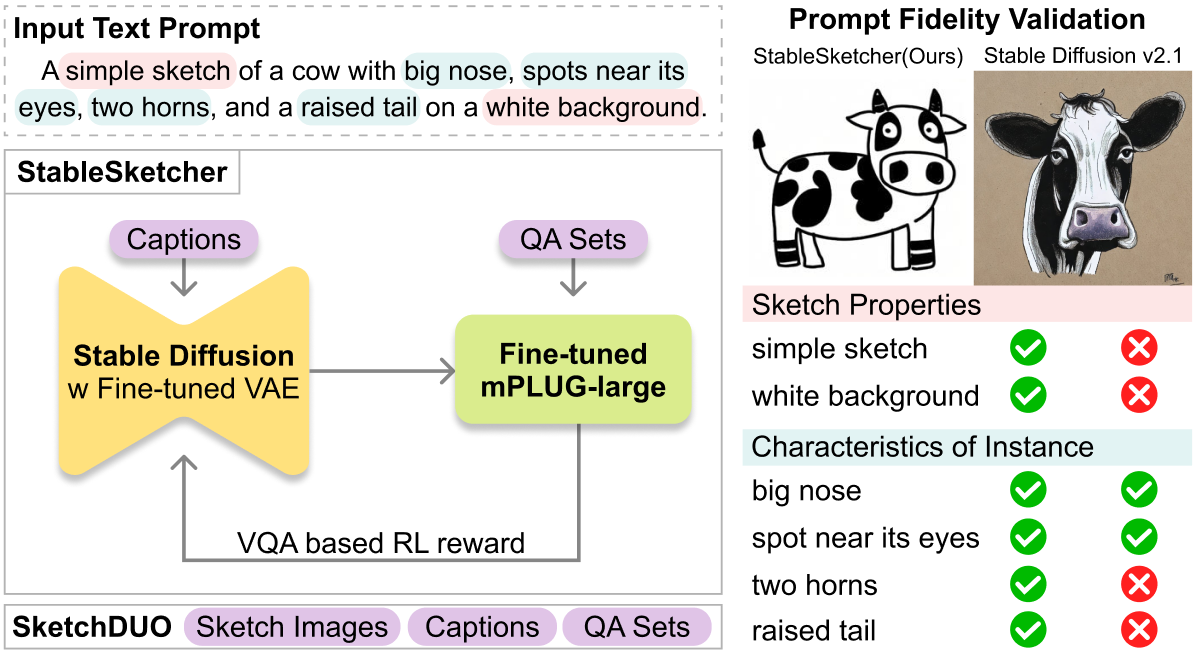}
    \caption{An overview of our StableSketcher framework and SketchDUO dataset. 
    % StableSketcher enhances Stable Diffusion to generate sketch images with high prompt fidelity through VAE fine-tuning and VQA-based RL feedback. SketchDUO is a dataset consisting of sketch images paired with captions and question-answer sets.
    }
    \label{fig:teaser}
    \vspace{-15pt}
\end{figure}

%% file: tex/2_related_works.tex
\section{Literature Review}

\begin{figure*}[t]
\centering
\includegraphics[width=\linewidth]{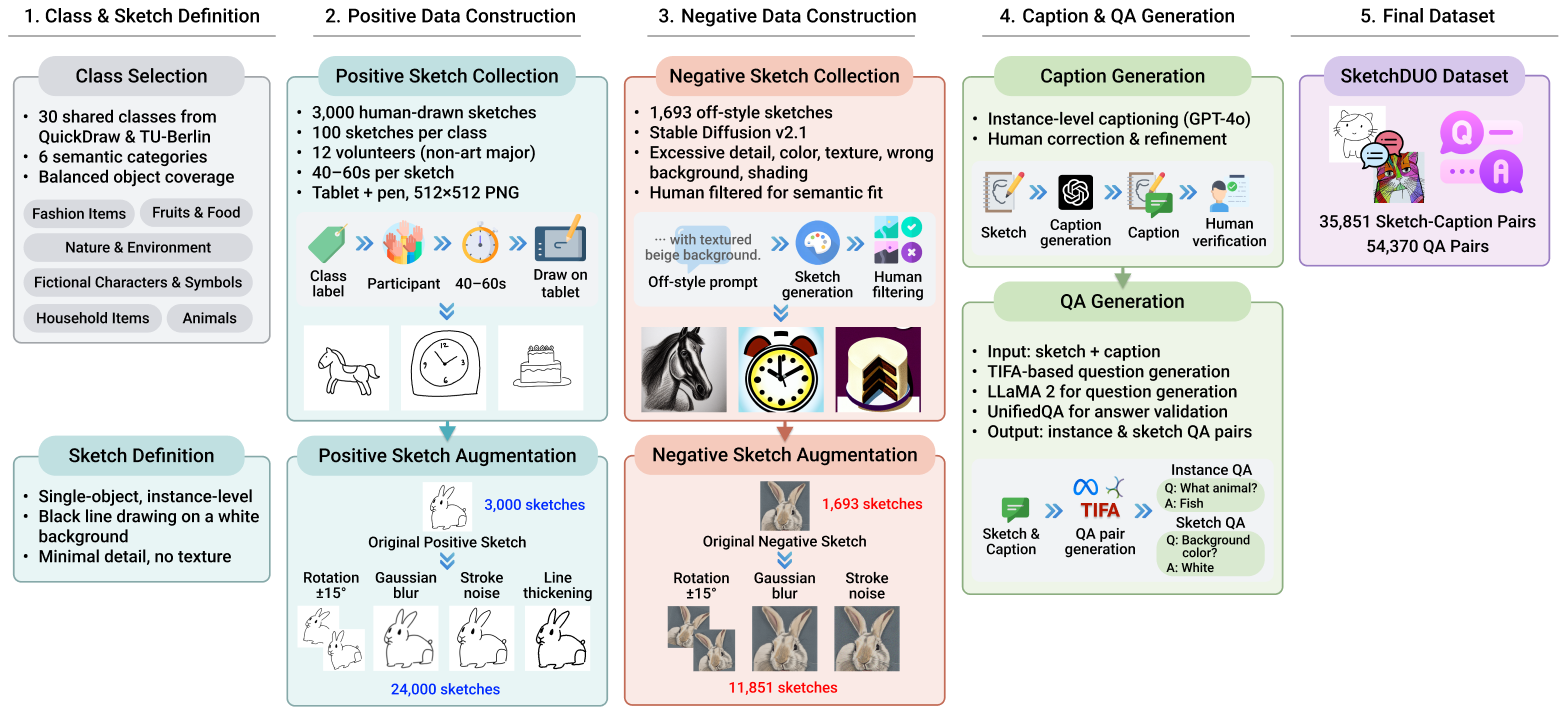}
\caption{SketchDUO construction process. The pipeline proceeds from left to right. It begins with class selection and sketch definition, followed by positive sketch collection and negative sketch construction. The two subsets are then augmented. In the caption generation stage, GPT-4o produces template-constrained instance-level captions for positive and negative sketches, and these captions are manually refined before TIFA-based QA pairs are constructed from each sketch--caption pair. The final SketchDUO dataset contains 35,851 sketch--caption pairs and 54,370 QA pairs.}
\label{fig:data_construction}
\end{figure*}
\renewcommand{\arraystretch}{1.15}

\setcellgapes{2.5pt}
\makegapedcells
\newcommand{\qapair}[2]{\parbox[t]{\linewidth}{Q.~#1\\A.~#2}}

\begin{table*}[t]
\centering
\footnotesize
\setlength{\tabcolsep}{4pt}
\renewcommand{\arraystretch}{1.12}
\caption{Representative examples from SketchDUO, showing positive--negative pairs for two object classes: fish (top two rows) and cake (bottom two rows).}
\label{tab:data-ex}
\begin{tabular}{p{0.12\textwidth} p{0.16\textwidth} p{0.18\textwidth} p{0.16\textwidth} p{0.28\textwidth}}
\hline

\multirow{4}{*}{\centering\raisebox{-0.5\height}{\includegraphics[width=0.10\textwidth]{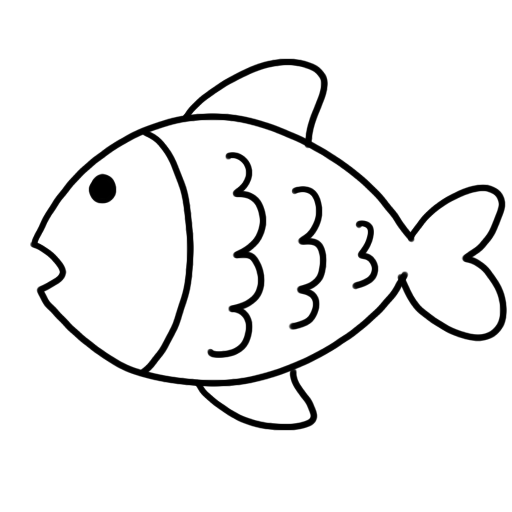}}}
& \textbf{Data Type} & \makecell[tl]{positive} & \textbf{File name} & \makecell[tl]{fish\_49.png} \\ \cline{2-5}
& \textbf{Caption} & \multicolumn{3}{p{0.62\textwidth}}{\parbox[t]{0.62\textwidth}{A simple drawing of a fish with three curved lines on its body and a round eye on a white background.}} \\ \cline{2-5}
& \makecell[tl]{\textbf{Instance-related}\\\textbf{Question--Answer}}
& \qapair{What animal is in the picture?}{Fish}
& \multicolumn{2}{p{0.44\textwidth}}{\qapair{How many lines are on the fish?}{3}} \\ \cline{2-5}
& \makecell[tl]{\textbf{Sketch-related}\\\textbf{Question--Answer}}
& \qapair{Is the background white?}{Yes}
& \multicolumn{2}{p{0.44\textwidth}}{\qapair{Is this a simple or a complex drawing?}{Simple}} \\ \hline

\multirow{4}{*}{\centering\raisebox{-0.5\height}{\includegraphics[width=0.10\textwidth]{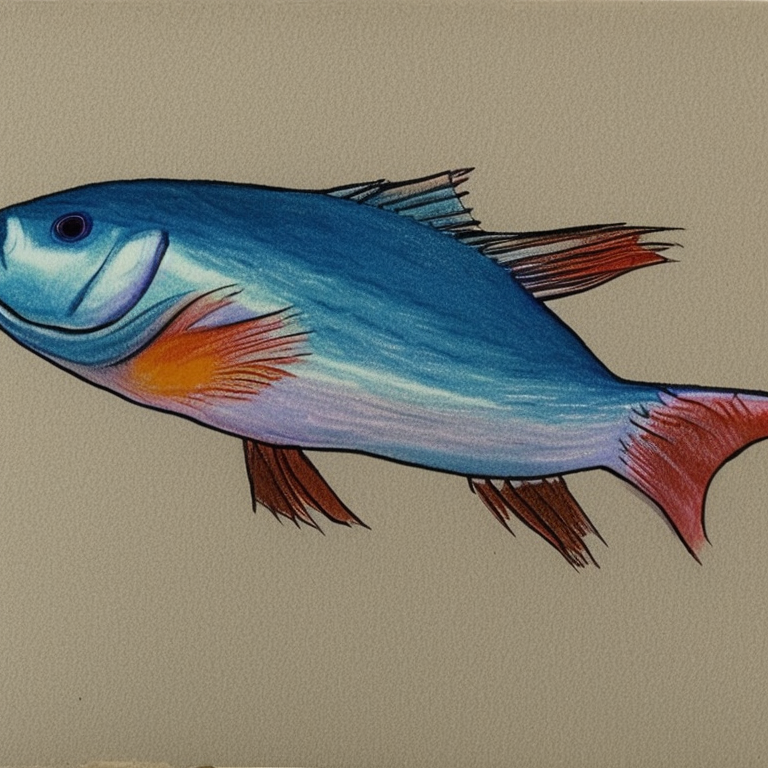}}}
& \textbf{Data Type} & \makecell[tl]{negative} & \textbf{File name} & \makecell[tl]{fish\_2.png} \\ \cline{2-5}
& \textbf{Caption} & \multicolumn{3}{p{0.62\textwidth}}{\parbox[t]{0.62\textwidth}{A detailed drawing of a blue and red fish with orange accents\\on a beige background featuring a lot of shading.}} \\ \cline{2-5}
& \makecell[tl]{\textbf{Instance-related}\\\textbf{Question--Answer}}
& \qapair{What color is the fish?}{Blue and red}
& \multicolumn{2}{p{0.44\textwidth}}{\qapair{How many lines are on the fish?}{3}} \\ \cline{2-5}
& \makecell[tl]{\textbf{Sketch-related}\\\textbf{Question--Answer}}
& \qapair{Is there a detailed drawing?}{Yes}
& \multicolumn{2}{p{0.44\textwidth}}{\qapair{Is there a lot of or a little shading?}{A lot of}} \\ \hline\hline

\multirow{4}{*}{\centering\raisebox{-0.5\height}{\includegraphics[width=0.10\textwidth]{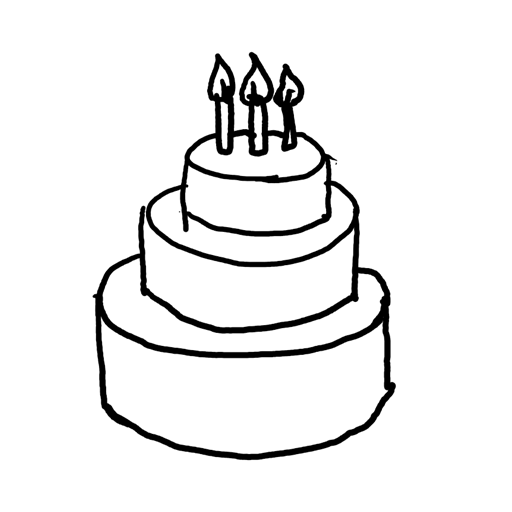}}}
& \textbf{Data Type} & \makecell[tl]{positive} & \textbf{File name} & \makecell[tl]{cake\_26.png} \\ \cline{2-5}
& \textbf{Caption} & \multicolumn{3}{p{0.62\textwidth}}{\parbox[t]{0.62\textwidth}{A simple drawing of a three-tier cake with three candles on a white background.}} \\ \cline{2-5}
& \makecell[tl]{\textbf{Instance-related}\\\textbf{Question--Answer}}
& \qapair{What is on the cake?}{Candles}
& \multicolumn{2}{p{0.44\textwidth}}{\qapair{Is the cake three-tier or two-tier?}{Three-tier}} \\ \cline{2-5}
& \makecell[tl]{\textbf{Sketch-related}\\\textbf{Question--Answer}}
& \qapair{What color is the background?}{White}
& \multicolumn{2}{p{0.44\textwidth}}{\qapair{Is this a simple or a complex drawing?}{Simple}} \\ \hline

\multirow{4}{*}{\centering\raisebox{-0.5\height}{\includegraphics[width=0.10\textwidth]{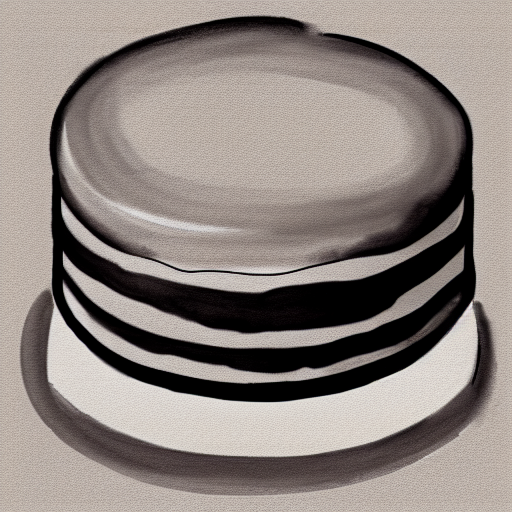}}}
& \textbf{Data Type} & \makecell[tl]{negative} & \textbf{File name} & \makecell[tl]{cake\_47.png} \\ \cline{2-5}
& \textbf{Caption} & \multicolumn{3}{p{0.62\textwidth}}{\parbox[t]{0.62\textwidth}{A layered cake with alternating dark and light layers on a beige background featuring a lot of shading.}} \\ \cline{2-5}
& \makecell[tl]{\textbf{Instance-related}\\\textbf{Question--Answer}}
& \qapair{What type of food is this?}{Layered cake}
& \multicolumn{2}{p{0.44\textwidth}}{\qapair{What color are the layers?}{Dark and light}} \\ \cline{2-5}
& \makecell[tl]{\textbf{Sketch-related}\\\textbf{Question--Answer}}
& \qapair{What is the background color?}{Beige}
& \multicolumn{2}{p{0.44\textwidth}}{\qapair{Is there a lot of shading?}{Yes}} \\ \hline

\end{tabular}
\end{table*}

\begin{figure*}[h!]
    \centering
    \includegraphics[width=0.85\linewidth, keepaspectratio]{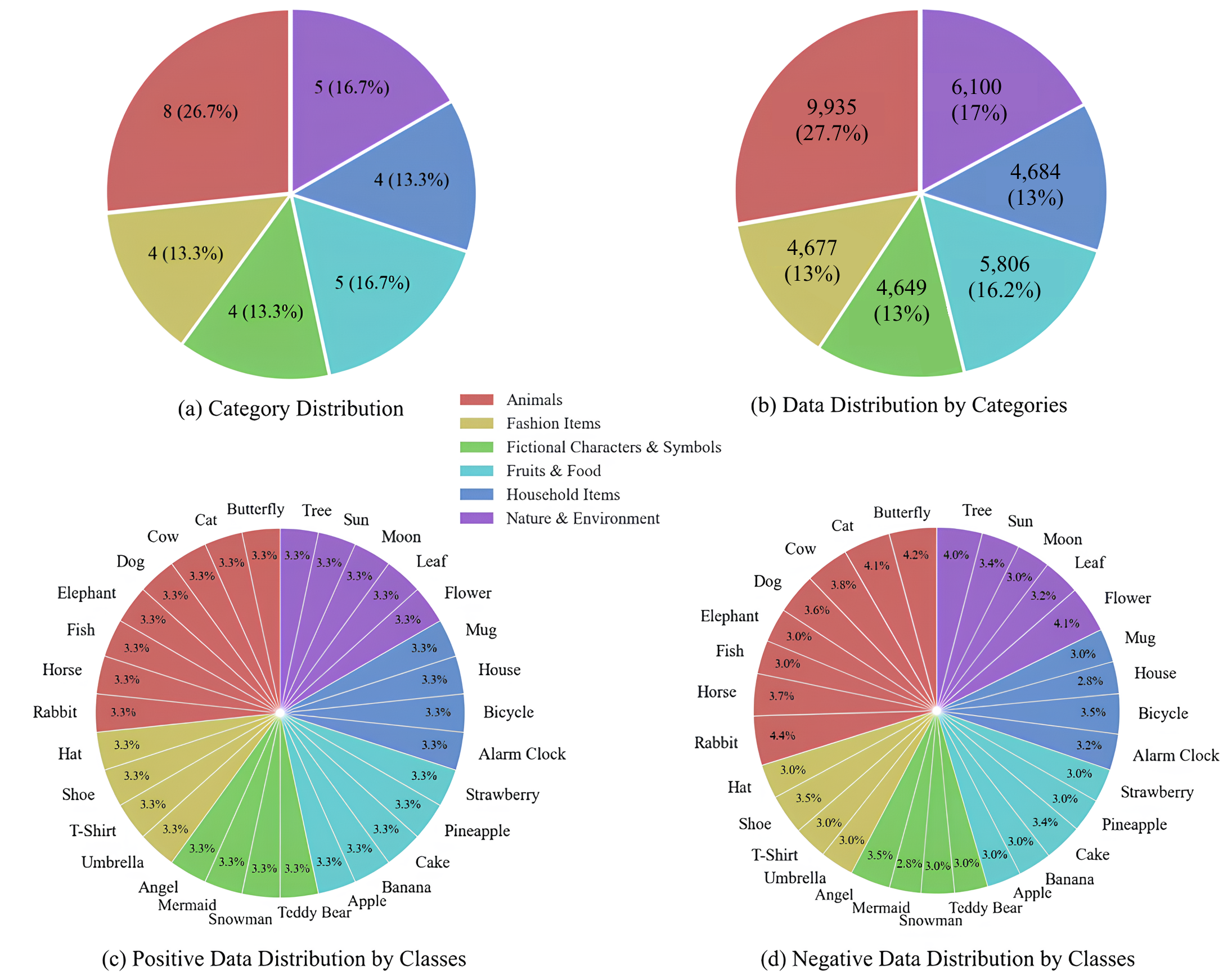}
    \caption{(a) Proportional distribution of the six categories in SketchDUO, shown as the number of categories and their respective percentages.
(b) Number of data samples within each category in SketchDUO, displayed as counts and their respective percentages.
(c) Class-level percentage distribution in the positive dataset.
(d) Class-level percentage distribution in the negative dataset.}
    \label{fig:pies}
\end{figure*}

\begin{figure}[t]
    \centering
    \includegraphics[width=\linewidth]{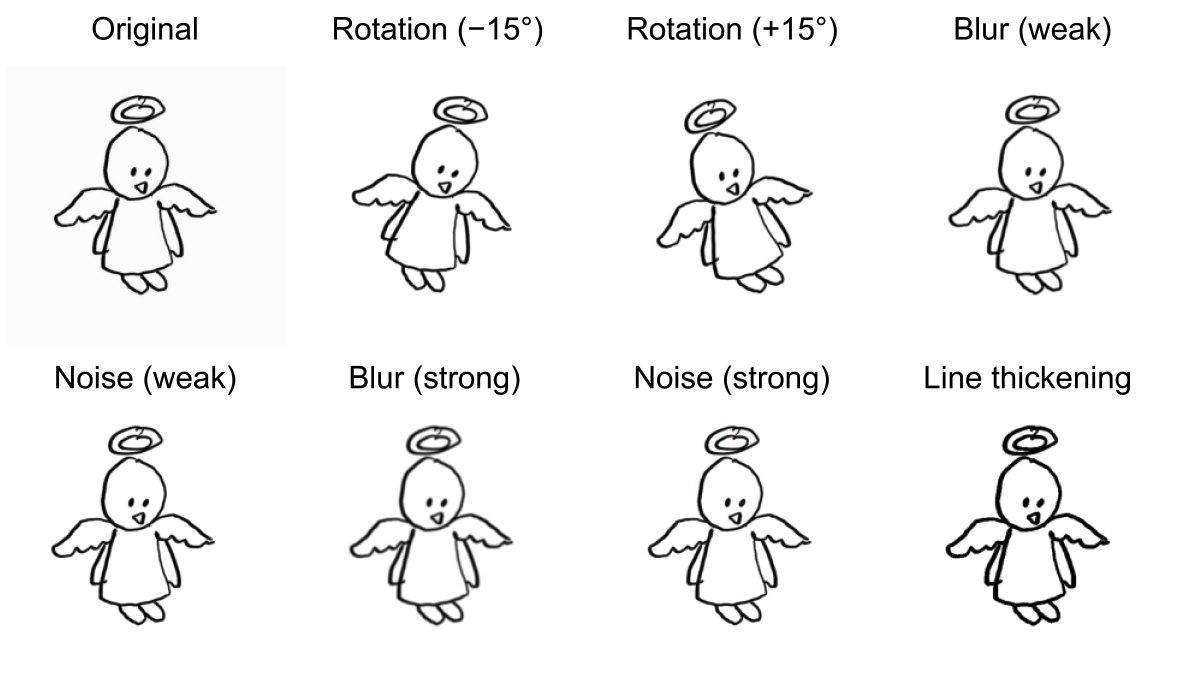}
    \caption{Illustration of background-preserving sketch augmentations. The top row, from left to right, shows the original sketch, rotation by $-15^\circ$ with white padding, rotation by $+15^\circ$ with white padding, and weak Gaussian blur ($k=3$, $\sigma=0.8$). The bottom row presents weak Gaussian noise applied only to the stroke regions ($\sigma=8$), strong Gaussian blur ($k=5$, $\sigma=1.6$), strong Gaussian noise applied to the stroke regions ($\sigma=16$), and line thickening obtained through binarization followed by morphological dilation with a $3\times3$ elliptical kernel for one iteration.}
    \vspace{-10pt}
\label{fig:sketch-aug}
\end{figure}

In this section, we first review the literature on \emph{sketch generation with diffusion models}, followed by a discussion on \emph{sketch datasets} and the application of \emph{reinforcement learning in diffusion models}.

\subsection{Sketch Generation with Diffusion Models}
Diffusion models are typically trained on large datasets of photorealistic images, resulting in a bias towards generating realistic, highly detailed outputs. This training bias limits their ability to generate abstract representations, such as sketches~\cite{zhan2025coprosketch,li2025text,yue2024diffusion,koley2024s}. Furthermore, conventional diffusion models often lack fine-grained control over the structural and abstract elements in sketches, making it difficult to achieve the desired level of simplicity and abstraction~\cite{koley2024s,wang2023diffsketching}. Although strategies to bridge the specific domain gap between sketches and photos have been extensively explored in heterogeneous face recognition through normalization and disentanglement frameworks~\cite{pang2024heterogeneous,pang2025unified}, effectively adapting these insights to the generative domain remains an open challenge.

At the same time, much of the sketch generation literature has focused on vector- and stroke-based representations, which model sketches at the granularity of individual strokes~\cite{ha2018neural,vinker2022clipasso}. While these approaches offer computational efficiency and editability, they often struggle to capture more complex and detailed sketches in a natural human-drawn style. Recent diffusion-based research has explored sketch synthesis in multiple representation spaces. For example, SwiftSketch~\cite{arar2025swiftsketch} proposes an image-conditioned diffusion model for vector sketch generation, enabling efficient synthesis by progressively denoising stroke control points. In parallel, Hu et al.~\cite{hu2024scaleadaptive} extend diffusion models to pixel-level sketch generation through a scale-adaptive sampling strategy for complex sketch synthesis.

More recent work has also studied controllable diffusion-based sketch generation using continuous sketch representations, such as unsigned distance fields, to improve structural clarity and controllability~\cite{zhan2025coprosketch}. However, these methods either focus on image-conditioned vector synthesis or depend on representations and training settings that do not directly address pixel-based generation of human-drawn sketches with fine-grained textual fidelity.

\subsection{Sketch Datasets}
As research in generative models progresses, a growing variety of sketch datasets has emerged to support advancements in sketch-related studies. QuickDraw~\cite{ha2018neural} is one of the largest datasets for sketch classification, but the absence of annotations and low-quality sketches limit its applicability to generative tasks. TU-Berlin~\cite{eitz2012hdhso} provides more complex sketches, but it also lacks descriptions of individual instances. Sketchy~\cite{sangkloy2016sketchy} pairs sketches with images, but the complexity of the sketches and the reliance on image-label pairs limit its usability for sketch generation. More recent efforts, such as SEVA~\cite{mukherjee2024seva}, leverage CLIPasso~\cite{vinker2022clipasso} to construct stroke-based sketch data for evaluating abstraction, yet they still rely on photo-sketch pairs and do not provide the fine-grained, instance-level captions needed for high-quality text-to-sketch generation. Likewise, SketchRef~\cite{lin2025sketchref} emphasizes benchmark construction and evaluation metrics for automated sketch synthesis, focusing on recognizability, structural consistency, and fairness across different simplification levels rather than providing richly annotated instance-level sketch--caption pairs for generative training. Therefore, existing datasets remain limited for learning pixel-based sketch generation with fine-grained semantic alignment.

\subsection{Reinforcement Learning in Diffusion Models}
The integration of reinforcement learning (RL) with diffusion models has recently attracted increasing attention, particularly for text-to-image alignment. Early work explored reward-weighted regression (RWR) to align generated images with textual prompts~\cite{lee2023aligning}, but such reward-weighted objectives can suffer from instability and limited credit assignment. More direct RL formulations have therefore been proposed for diffusion fine-tuning. For example, diffusion policy optimization with KL regularization (DPOK)~\cite{fan2023dpok} formulates text-to-image diffusion fine-tuning as an online RL problem with policy gradient optimization and KL regularization. Building on this perspective, denoising diffusion policy optimization (DDPO)~\cite{black2023training} models the denoising process itself as a sequential decision-making problem, allowing diffusion models to be optimized directly with user-defined reward functions. More recent work has further extended RL-based diffusion training to address sparse-reward settings during denoising~\cite{hu2025towards}. 

DDPO is particularly relevant to our setting because it enables direct optimization with task-specific black-box rewards defined on generated images. In the original DDPO framework, BERTScore~\cite{zhang2019bertscore} is used as a text-image alignment reward, but such caption-based similarity is limited for abstract forms like sketches. We instead leverage the TIFA score~\cite{hu2023tifa}, which uses question-answer based evaluation for a more fine-grained assessment of text-image alignment. Building on TIFA, we devise a novel VQA-based reward function and incorporate it into DDPO training to improve the semantic fidelity of generated sketches.

%% file: tex/3_dataset.tex
\section{SketchDUO}

SketchDUO contains 35,851 instance-level sketch images paired with textual captions and 54,370 QA sets. By offering both captions and QA pairs, SketchDUO provides rich descriptions for individual objects, effectively addressing the limitations of existing datasets. The dataset adopts a contrastive approach, featuring positive examples that capture the desired sketch style, and negative examples that highlight common misrepresentations in Stable Diffusion outputs, such as images with excessive detail, over-shading, or sketches that resemble photographs of pencil drawings rather than true hand-drawn representations. SketchDUO was constructed through a human-in-the-loop pipeline, in which model-generated captions and QA candidates were subsequently verified and revised by human annotators to ensure semantic accuracy and subset-specific stylistic consistency.

Figure~\ref{fig:data_construction} provides an overview of the SketchDUO construction pipeline, and Table~\ref{tab:data-ex} presents example images, captions, and QA sets from SketchDUO.
To construct SketchDUO, we selected 30 common classes shared between the QuickDraw~\cite{ha2018neural} and TU-Berlin~\cite{eitz2012hdhso} datasets. The selection of classes was designed to achieve a balanced representation of diverse objects, ensuring broad thematic coverage across the dataset. SketchDUO comprises 30 classes, distributed across six broad categories to ensure balanced representation and thematic diversity. Below, we outline the six main categories and their corresponding classes:

\begin{enumerate}
    \item \textbf{Fashion Items:} Hat, Shoe, T-shirt, Umbrella.
    \item \textbf{Animals:} Butterfly, Cat, Cow, Dog, Elephant, Fish, Horse, Rabbit.
    \item \textbf{Nature \& Environment:} Flower, Leaf, Moon, Sun, Tree.
    \item \textbf{Fictional Characters \& Symbols:} Angel, Mermaid, Snowman, Teddy Bear
    \item \textbf{Fruits \& Food:} Apple, Banana, Cake, Pineapple, Strawberry.
    \item \textbf{Household Items:} Alarm Clock, Bicycle, House, Mug.
    
\end{enumerate}

Figure~\ref{fig:pies} visualizes the category and class distributions, including proportions across categories, class allocations, and sample distributions in the positive and negative datasets.

\subsection{Definition of a Sketch}
We define a sketch as a simple, human-drawn representation of a single instance. The sketch is characterized by a black line drawing on a white background with no texture, capturing the essence of the object with minimal complexity. The representation must be instance-level, focusing on a single object without excessive details or unnecessary elements.

\subsection{Sketch Image Collection}
\begin{table*}[t]
\centering
\footnotesize
\caption{GPT-4o captioning prompts and human revision criteria used in SketchDUO construction. The positive and negative datasets use different prompt templates to reflect the target sketch style and controlled off-style contrastive examples, respectively.}
\label{tab:caption-prompts}

\setlength{\fboxsep}{6pt}
\setlength{\fboxrule}{0.5pt}

\fbox{%
\parbox[t]{0.46\textwidth}{%
\raggedright
\textbf{Positive dataset}\par
\vspace{-0.5em}
\rule{\linewidth}{0.4pt}\par
% \vspace{0.4em}

\textbf{Model:} GPT-4o\par
\textbf{System message:} You are a helpful assistant for generating image captions.\par
\textbf{Input:} An image provided as a base64-encoded JPEG string.\par

\vspace{0.5em}
\textbf{Output template}\par
\hspace*{1em}``A black line drawing of \{\{text1\}\} on a white background.''\par
\hspace*{1em}\textbf{or}\par
\hspace*{1em}``A simple drawing of \{\{text1\}\} on a white background.''\par

\vspace{0.5em}
\textbf{Key prompt constraints}\par
\hspace*{1em}-- Replace \{\{text1\}\} with a detailed instance-level description.\par
\hspace*{1em}-- Focus on clear objects, shapes, and actions rather than vague category-only descriptions.\par
\hspace*{1em}-- Choose ``black line drawing'' for relatively detailed sketches and ``simple drawing'' for more minimal sketches.\par
\hspace*{1em}-- Do not output the placeholder brackets.\par

\vspace{0.5em}
\textbf{Human revision criteria}\par
\hspace*{1em}-- Correct inaccurate object attributes, counts, and local structural details.\par
\hspace*{1em}-- Remove vague wording.\par
\hspace*{1em}-- Ensure consistency with the positive sketch definition: a single-object black-line drawing on a white background with minimal detail.\par
}%
}%
\hfill
\fbox{%
\parbox[t]{0.47\textwidth}{%
\raggedright
\textbf{Negative dataset}\par
\vspace{-0.5em}
\rule{\linewidth}{0.4pt}\par
% \vspace{0.4em}

\textbf{Model:} GPT-4o\par
\textbf{System message:} You are a helpful assistant for generating image captions.\par
\textbf{Input:} An image provided as a base64-encoded JPEG string.\par

\vspace{0.5em}
\textbf{Output template}\par
\hspace*{1em}``A \{\{text1\}\} drawing of \{\{text2\}\} on a \{\{text3\}\} background \{\{text4\}\}.''\par

\vspace{0.5em}
\textbf{Key prompt constraints}\par
\hspace*{1em}-- Replace \{\{text1\}\} with ``detailed'' or ``simple'' according to sketch complexity.\par
\hspace*{1em}-- Replace \{\{text2\}\} with a detailed object description.\par
\hspace*{1em}-- Replace \{\{text3\}\} with the background color.\par
\hspace*{1em}-- Replace \{\{text4\}\} with ``featuring a lot of shading.'' when shading is present; otherwise leave it blank.\par
\hspace*{1em}-- This template explicitly allows off-style traits such as color, non-white background, and shading.\par

\vspace{0.5em}
\textbf{Human revision criteria}\par
\hspace*{1em}-- Correct semantic errors while preserving contrastive off-style properties.\par
\hspace*{1em}-- Retain color, excessive detail, non-white background, and shading when visually present.\par
\hspace*{1em}-- Preserve the negative subset's role as controlled off-style supervision.\par
}%
}%

\end{table*}
We curate a corpus of $24,000$ positive and $11,851$ negative sketch images. The positive portion is derived from $3,000$ human-drawn sketches spanning 30 classes, with one hundred sketches per class. The negative portion contains $1,693$ images generated with Stable Diffusion v2.1 and selected for off style traits such as intricate line work, the presence of color, nonwhite backgrounds, and heavy shading. 

To every image we apply the same seven background-preserving augmentations. For negative samples, which contain no strokes, we omit the line thickening augmentation. These include rotations of plus or minus fifteen degrees with white padding; Gaussian blur with a weak setting where $k$ equals three and sigma equals $0.8$, and a strong setting where $k$ equals five and $\sigma$ equals 1.6; Gaussian noise applied to strokes only with a weak setting where $\sigma$ equals eight and a strong setting where $\sigma$ equals sixteen; and line thickening achieved through morphological dilation, as detailed in Figure~\ref{fig:sketch-aug}. The resulting collection contains $35,851$ images in total, comprising $24,000$ positive and $11,851$ negative samples. A comprehensive analysis of the distributions at the category and class levels for both subsets, including relative proportions and sample counts, is presented in Figure~\ref{fig:pies}.

\subsubsection{Positive Sketch Data Construction}
To construct the positive samples, we collected 3{,}000 hand-drawn sketch images, with 100 instances per class. These sketches were created by 12 volunteer participants, none of whom were art majors. All participants followed a standardized drawing protocol to ensure stylistic and semantic consistency across the dataset. They were provided with reference examples illustrating the desired visual characteristics of positive sketches. All sketches were drawn using a tablet and pen with the default drawing tool to maintain consistency in the drawing tools and procedures. The images were saved in PNG format at a fixed resolution of 512{\(\times\)}512 pixels. The drawing style was carefully controlled to reflect the core properties of the dataset. Each sketch was required to depict a single object instance in a minimal style using thick black lines (\#000000) on a pure white background (\#FFFFFF). Participants were instructed to complete each drawing within 40 to 60 seconds to encourage simplicity and consistency in visual abstraction.

Captions for each image were initially generated with GPT-4o using a template-constrained prompt tailored to the positive sketches. The prompt restricted the output to a single-object black-on-white sketch description, consistent with our sketch definition. Each generated caption was then manually revised to correct object attributes, counts, and local structural details before QA construction. The exact prompt is provided in Table~\ref{tab:caption-prompts}. 

These finalized captions were then passed to a question generation module adapted from the TIFA framework~\cite{hu2023tifa}, which leverages LLaMA 2~\cite{touvron2023llama} for generating diverse questions and UnifiedQA~\cite{khashabi2020unifiedqa} for answer validation. This process yielded structured QA sets for each image, ensuring that every positive sample in the dataset is paired with high-quality captions and reliable question–answer triplets.

\subsubsection{Negative Sketch Data Construction}
To build a high-quality negative sample set for contrastive learning, we constructed 1{,}693 sketch--caption--QA triplets. All images were generated using Stable Diffusion v2.1, and captions were produced with GPT-4o using a separate prompt tailored to off-style sketches. Unlike the positive template, this prompt explicitly allowed descriptions of non-white backgrounds, color, excessive detail, and shading, which are central attributes of the negative subset. The generated captions were then manually revised to preserve semantic correctness while retaining the undesired stylistic traits needed for contrastive supervision. The exact prompt is provided in Table~\ref{tab:caption-prompts}.

QA pairs were generated using the TIFA framework~\cite{hu2023tifa}, which combines LLaMA 2~\cite{touvron2023llama} for question generation and UnifiedQA~\cite{khashabi2020unifiedqa} for answer validation. Negative samples were generated through a systematic procedure focused on producing images that diverge from the target sketch style. These controlled stylistic deviations serve as contrastive supervision, enabling the model to better distinguish desired sketch attributes from undesired ones. The prompts were explicitly designed to generate characteristics inconsistent with the black-on-white sketch aesthetic, including detailed line drawings, colored elements, textured or non-white backgrounds, and heavy shading. Following generation, all candidate images underwent human filtering to ensure inclusion criteria and semantic fidelity.

\subsection{Sketch-Caption Pair}
\label{sec:sketch-caption-pair}

To construct high-quality sketch--caption pairs, we initially evaluated several image captioning models, including BLIP-2 Flan T5-xl~\cite{li2023blip}, mPLUG~\cite{li2022mplug}, mPLUG-Owl3~\cite{ye2024mplug}, and GPT-4o~\cite{hurst2024gpt}. Smaller captioning models often produced generic, incorrect, or overly simplified descriptions for sketches, especially for instance-specific attributes such as counts, local shapes, and background conditions. We therefore adopted GPT-4o as the primary captioning model and used template-constrained prompts tailored to SketchDUO. This caption construction process followed a human-in-the-loop pipeline, where GPT-4o first generated candidate captions and human annotators then verified and corrected them before QA generation.

As summarized in Table~\ref{tab:caption-prompts}, the positive prompt constrained the description to a black-on-white single-object sketch format, whereas the negative prompt explicitly allowed off-style properties such as non-white backgrounds, color, and heavy shading. This prompt design was intended to reflect the distinct roles of the two subsets in the contrastive dataset construction pipeline.

All GPT-4o captions were then manually revised before QA generation. Human post-editing focused on correcting object identity, counts, local attributes, and background descriptions while preserving the intended subset-specific style. For positive sketches, annotators normalized the captions to remain faithful to the black-line, white-background sketch definition. For negative sketches, annotators preserved semantically correct object descriptions while explicitly retaining undesired stylistic traits such as color, excessive detail, non-white background, and shading.

\subsection{Sketch-QA Triplet}
QA sets are generated using the question generation module from the TIFA framework~\cite{hu2023tifa}, which combines LLaMA 2~\cite{touvron2023llama} for question generation and UnifiedQA~\cite{khashabi2020unifiedqa} for validating the generated questions. The dataset comprises both positive and negative triplets, with each triplet consisting of a sketch, a corresponding question, and its answer. The positive dataset contains 37,412 QA pairs, while the negative dataset includes 16,958 QA pairs, resulting in a total of 54,370 Sketch-QA triplets. These triplets are crafted to provide rich semantic detail and understanding of single-object sketches.

%% file: tex/4_method.tex
\section{StableSketcher}
\begin{figure*}[t]
\centering
\includegraphics[width=0.9\linewidth]{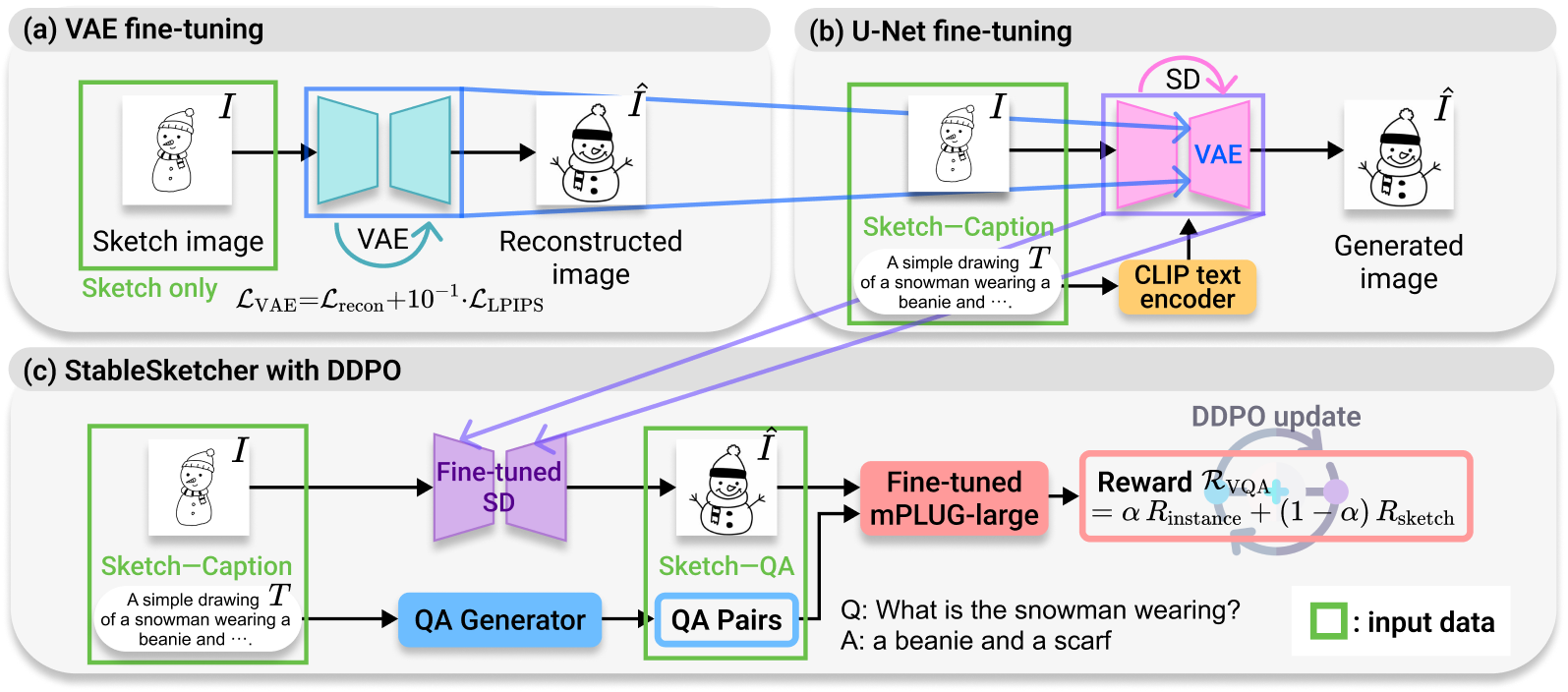}
\caption{Overall architecture of StableSketcher. The input prompt ($T$) is fed into the diffusion model through the CLIP text encoder, where the VAE is fine-tuned using $\mathcal{L}_{\text{recon}}$ and $\mathcal{L}_{\text{LPIPS}}$. Once the diffusion model generates the image ($\hat{I}$), it is passed to the fine-tuned mPLUG-large, along with the question from SketchDUO, to generate the corresponding answer. The VQA-based reward $\mathcal{R}_{\text{VQA}}$, calculated using the TIFAScore, is then used as a feedback signal in the reinforcement learning process.}
\label{fig:framework}
\end{figure*}
\subsection{Backgrounds}

\subsubsection{Diffusion Models}
Diffusion models are generative models that synthesize data by reversing a gradual noising process~\cite{ho2020denoising}. Starting from a clean sample $x_0$, Gaussian noise is incrementally added through the forward process $q(x_t|x_{t-1})$, until pure noise is reached at step $T$. The model then learns a reverse denoising process $p_\theta(x_{t-1}|x_t)$ to reconstruct data from noise. A common training objective is the noise prediction (score-matching) loss:
\begin{equation}
\mathcal{L}_{DM}(\theta) = \mathbb{E}_{x_0,\epsilon,t}\Big[\|\epsilon - \epsilon_\theta(x_t,t)\|^2\Big],
\end{equation}
where $\epsilon_\theta$ predicts the noise $\epsilon$ added at step $t$. By chaining this reverse process, diffusion models can sample high-quality and diverse outputs from pure noise.

\subsubsection{Latent Diffusion Models (LDMs)}
Stable Diffusion~\cite{rombach2022high} adapts this framework into a latent space for efficiency. Instead of operating directly in pixel space, an image $x_0$ is encoded into a latent representation $z$ via a VAE, and the diffusion process is carried out in this lower-dimensional space. The training objective then becomes:
\begin{equation}
\mathcal{L}_{LDM}(\theta) = \mathbb{E}_{z,\epsilon,t,c}\Big[\|\epsilon - \epsilon_\theta(z_t, t, c)\|^2\Big],
\end{equation}
where $c$ denotes conditioning information such as a textual prompt. By performing diffusion in latent space, LDMs significantly reduce computational cost while maintaining the ability to generate semantically aligned images conditioned on text.

\subsubsection{Denoising Diffusion Policy Optimization}
While diffusion models generate visually realistic images, they may not align closely with input conditions (e.g., textual prompts) or task-specific objectives. DDPO ~\cite{black2023training} addresses this limitation by framing the denoising process as a Markov decision process (MDP)~\cite{sutton2018reinforcement}, where each denoising step is treated as an action. At timestep $t$, the model predicts a denoised sample $x_{t-1}$ and receives a reward $r(x_{t-1}, x_0, y)$ measuring alignment with the conditioning input $y$. The optimization objective is to maximize the expected cumulative reward:
\begin{equation}
\max_\theta \; J(\theta) = \mathbb{E}_{x_0,y}\Bigg[\sum_{t=1}^T r\big(x_{t-1}, x_0, y\big)\Bigg].
\end{equation}

This is optimized using a policy-gradient method adapted to diffusion models. The policy is defined as $\pi_\theta(x_{t-1}|x_t, t, y)$, and the gradient is estimated in REINFORCE style:
\begin{equation}
\nabla_\theta J(\theta) \approx \mathbb{E}\Bigg[\sum_{t=1}^T \nabla_\theta \log \pi_\theta(x_{t-1}|x_t, t, y)\, R_t \Bigg],
\end{equation}
where $R_t$ denotes the cumulative return. This framework enables diffusion models to go beyond maximum likelihood training by directly incorporating task-specific feedback, such as prompt fidelity, stylistic constraints, or user-defined rewards. In our work, we employ DDPO with a VQA-based reward function to explicitly improve the semantic alignment of generated sketches with their textual prompts.

Building upon these foundations, we propose \emph{StableSketcher}, a training framework that adapts Stable Diffusion for sketch generation and incorporates DDPO with VQA-based feedback. As illustrated in Figure~\ref{fig:framework}, StableSketcher takes as input a textual prompt $\mathbf{T}$ and a human-drawn sketch $\mathbf{I}$ from the SketchDUO dataset. First, the VAE is fine-tuned on sketch images to enhance reconstruction. Next, the U-Net of Stable Diffusion is trained with the adapted VAE to generate human-drawn style sketches conditioned on $\mathbf{T}$. Finally, Stable Diffusion is fine-tuned with DDPO using our VQA-based reward, which extracts elements from $\mathbf{T}$ and evaluates them individually to improve prompt fidelity. The training procedure can be divided into two main stages: (i) Training Stable Diffusion and (ii) VQA Feedback using DDPO, detailed in the following subsections.

\subsection{Training Stable Diffusion}
\subsubsection{VAE Fine-tuning for Sketch Reconstruction}
The Autoencoder KL \cite{kingma2013auto}, used as the frozen VAE in Stable Diffusion, has a loss function composed of two main components. 
First, the reconstruction loss $\mathcal{L}_{\text{recon}}$ measures how well the input data $x$ has been reconstructed via mean squared error (MSE). This can be expressed as:
\begin{equation}
    \mathcal{L}_{\text{recon}} = \| x - \hat{x} \|^2
    \label{eq:recon}
\end{equation}
Second, the Kullback–Leibler Divergence (KL) loss $\mathcal{L}_{\text{KL}}$ evaluates how close the distribution sampled from the latent space is to a normal distribution $\mathcal{N}(0,I)$. A weighting factor $\beta$ is often applied to balance the reconstruction and KL terms:
\begin{align}
    \mathcal{L}_{\text{AutoencoderKL}} &= \mathcal{L}_{\text{recon}} + \beta \cdot \mathcal{L}_{\text{KL}} \label{eq:vae1} \\
    &= \| x - \hat{x} \|_2^2 + D_{\text{KL}}(q(z|x) \| p(z)) \label{eq:vae2}
\end{align}

Using a \emph{large} KL term over-regularizes the approximate posterior \(q_{\phi}(z \mid x)\) toward the standard normal prior, reducing the mutual information \(I(x;z)\) and causing posterior collapse, which leads to poor or even failed reconstructions. As shown in Figure~\ref{fig:kl_ablation}, collapse occurs when the KL weight is large, while reconstruction becomes feasible again when the weight is reduced to very small values. Conversely, relying solely on pixel-wise reconstruction loss, $L_{recon}$ can result in instability in the loss values, leading to unstable training. In particular, sketch data relies heavily on local and perceptual features such as contours and line thickness, which are difficult to capture with pixel-wise errors alone. Losses like MSE or KL do not adequately reflect these perceptual aspects.

To address this issue, we leverage learned perceptual image patch similarity (LPIPS)~\cite{zhang2018unreasonable} as a loss function to better capture the characteristics of sketches. LPIPS measures perceptual similarity based on multi-layer CNN feature maps, capturing not just pixel-level differences but also human-perceived properties such as line sharpness, shape consistency, and visual coherence. This makes it especially suitable for sketch images, where abstraction and contour fidelity are more critical than photorealistic detail. LPIPS is defined as:
\begin{equation}
    \mathcal{L}_{\text{LPIPS}} = \sum_{l} w_l \cdot \| \phi_l(x) - \phi_l(\hat{x}) \|^2,
\end{equation}
where $\phi_l(\cdot)$ denotes the feature map from the $l$-th layer.  

Therefore, our final training VAE loss combines MSE with LPIPS to achieve both stable training and sketch-specific reconstruction quality:
\begin{align}
    \mathcal{L}_{\text{VAE}} &= \| x - \hat{x} \|^2 + 10^{-1} \cdot \mathcal{L}_{\text{LPIPS}}.
\end{align}

\subsubsection{U-Net Fine-tuning for Text-Aligned Sketch Generation}
We perform U-Net fine-tuning on Stable Diffusion using sketch–caption pairs from the SketchDUO dataset to adapt the model for generating human-drawn style sketch images. As illustrated in Figure~\ref{fig:framework}, the frozen VAE is replaced with our enhanced VAE to better capture sketch-specific representations. Text prompts are incorporated into the U-Net through a cross-attention mechanism, enabling the model to effectively align the denoising process with the given prompt. Furthermore, the denoising diffusion probabilistic models (DDPM)~\cite{ho2020denoising} scheduler is employed to ensure a stable and consistent diffusion process during training. We follow the original noise prediction objective of Stable Diffusion~\cite{rombach2022high} for U-Net fine-tuning.

\subsection{VQA-Guided Fine-tuning with DDPO}

\subsubsection{Design of VQA-Based Reward Function}
DDPO~\cite{black2023training} originally employed BERTScore~\cite{zhang2019bertscore} with LLaVa~\cite{liu2024visual} to define a reward signal. However, BERTScore has limitations in capturing fine-grained representations, since it computes similarity based on captions generated by vision-language models (VLMs). In this process, the original image is first converted into a caption, which tends to preserve only coarse, overall semantics while discarding fine-grained visual details. As a result, BERTScore evaluates alignment at a global level but fails to verify whether individual elements of the prompt are accurately reflected in the generated image. To address this, we propose a new reward function inspired by TIFAScore~\cite{hu2023tifa}, which evaluates the prompt fidelity of text-to-image generation by checking whether each individual element of a text prompt is satisfied by the generated image. Formally, TIFAScore is defined as:
\begin{equation}
    \text{TIFAScore} = \frac{1}{N} \sum_{i=1}^{N} \delta(f(Q_i, I), A_i),
\end{equation}
where $N$ denotes the number of question and answer (QA) pairs, $Q_i$ is a question derived from the prompt, $I$ is the generated image, $f(\cdot)$ is a VQA model, $A_i$ is the ground-truth answer, and $\delta(\cdot)$ is the Kronecker delta function.

\subsubsection{Implementation VQA-Based Reward Function with SketchDUO QA Triplets}
Building on this idea, we design a reward function that captures both instance-level fidelity and sketch-style faithfulness using the sketch--QA triplets from SketchDUO:
\begin{equation}
\mathcal{R}_{\text{VQA}} = \alpha \cdot \mathcal{R}_{\text{instance}} + (1-\alpha) \cdot \mathcal{R}_{\text{sketch}}
\end{equation}

For each image, there are $N+M$ QA pairs, consisting of $N$ instance-related questions and $M$ sketch-style questions. The weighting ratio is controlled by $\alpha$, where $0 \leq \mathcal{R}_{\text{VQA}} \leq 1$. We set $\alpha = 0.5$ in our experiments:
\begin{align}
\mathcal{R}_{\text{instance}} &= \frac{1}{N} \sum_{i=1}^{N} \delta(f(Q^{\text{instance}}_i, I), A^{\text{instance}}_i), \\
\mathcal{R}_{\text{sketch}} &= \frac{1}{M} \sum_{j=1}^{M} \delta(f(Q^{\text{sketch}}_j, I), A^{\text{sketch}}_j).
\end{align}
For the VQA backbone, we adopt the mPLUG-large model~\cite{li2022mplug}, which achieves strong accuracy among SOTA VQA models with competitive inference time~\cite{hu2023tifa}.

This reward score $\mathcal{R}_{\text{VQA}}$ is used as the feedback signal in the DDPO training loop. At each training step, the sketch generation model produces candidate images based on text prompts, and $\mathcal{R}_{\text{VQA}}$ is computed by evaluating how well each generated image satisfies the paired questions. A higher $\mathcal{R}_{\text{VQA}}$ score indicates that the image successfully satisfies both semantic correctness and sketch-style intent. This reward guides the policy updates by reinforcing image generations that more faithfully reflect prompt semantics and human-like abstraction.

\input{tbl/vqa_acc}

\subsubsection{VQA Model Fine-tuning for Accurate Reward Signals}
The accuracy of the VQA model directly impacts the reliability of the reward signal and thus the quality of policy updates. High rewards are assigned when generated sketches align with prompt conditions, whereas mismatches are assigned lower rewards, guiding the policy toward faithful sketch generation. To improve sketch understanding, we fine-tune the VQA model on the SketchDUO QA set, using 80\% of the data for training and 20\% for evaluation. This fine-tuning uses both positive and negative sketch-QA triplets, allowing the model to learn not only whether a generated image matches the prompt semantics, but also whether it conforms to the intended sketch style. Consequently, the reward function can reflect both desired and undesired stylistic patterns. A comparison of the baseline mPLUG-large model and our fine-tuned model is presented in Table~\ref{tab:vqa_acc}.

%% file: tbl/vqa_acc.tex
\begin{table}[t]
\caption{Accuracy comparison of the mPLUG-Large VQA model fine-tuned with SketchDUO.}
\label{tab:vqa_acc}
\centering
\footnotesize
\renewcommand{\arraystretch}{1.2} % 행 간격 살짝 늘림
\setlength{\tabcolsep}{6pt}       % 열 간격 조정
\begin{tabular}{lccccc}
\toprule
\multirow{2}{*}{} & \multirow{2}{*}{\begin{tabular}[c]{@{}c@{}}Baseline\\ (mPLUG-L)\end{tabular}} & \multirow{2}{*}{Dataset} & \multicolumn{3}{c}{Fine-tuning (Epochs)} \\
\cmidrule{4-6}
 &  &  & 2 & 4 & 6 \\
\midrule
\multirow{2}{*}{Accuracy (\%)} & \multirow{2}{*}{61.38} & Positive & 87.38 & 88.39 & 88.80 \\
 &  & Both & 88.05 & 89.04 & \textbf{89.38} \\
\bottomrule
\end{tabular}
\end{table}

%% file: tex/5_experiments.tex
\section{Experiments}
\subsection{Implementation Details}
\subsubsection{Dataset: SketchDUO}
In this study, the sketch-image-caption dataset was divided into training and test sets using a 6:4 split, and each subset was used differently across learning stages. For VAE fine-tuning, sketch images from the training split were used for optimization, whereas the Stable Diffusion U-Net was trained on sketch-image-caption pairs from the same split. DDPO-based reinforcement learning was performed using training sketches and their associated QA pairs. In addition, fine-tuning of the mPLUG-large VQA model incorporated negative samples in undesired styles together with their corresponding captions and QA pairs.

\subsubsection{Training configuration}
Unless otherwise noted, all modules were trained on two NVIDIA RTX A6000 GPUs, and we report the main hyperparameters and approximate wall-clock training times for reproducibility. VAE fine-tuning was conducted for 15 epochs on a single NVIDIA RTX 3060 Ti GPU using AdamW (learning rate $1\times10^{-5}$, batch size 2, $\beta_1=0.9$, $\beta_2=0.999$, weight decay 0.01) with the VGG-based LPIPS and reconstruction losses; this stage took approximately 11 hours. U-Net fine-tuning then used SketchDUO at a resolution of $512\times512$ with center cropping and random horizontal flipping. We fine-tuned only the U-Net while freezing the VAE and CLIP text encoder, using AdamW (learning rate $1\times10^{-5}$, batch size 1), gradient accumulation of 1, gradient checkpointing, EMA, and a constant learning-rate scheduler without warmup for 100 epochs; this stage took approximately 30 hours.

DDPO-based fine-tuning started from our VAE-adapted Stable Diffusion checkpoint and optimized LoRA parameters using the TIFAScore-based reward. We trained for 800 epochs with batch size 1, gradient accumulation of 1, learning rate $3\times10^{-4}$, maximum gradient norm 1, and Adam ($\beta_1=0.9$, $\beta_2=0.999$, $\epsilon=10^{-8}$, weight decay $10^{-4}$), with FP16 mixed precision and TF32 enabled. Sampling used 50 denoising steps, guidance scale 5, $\eta=1$, and one sampled batch per epoch; this stage took approximately 16 hours. For VQA fine-tuning, we trained mPLUG-large on the SketchDUO QA set for 6 epochs using both positive and negative sketch--QA triplets, with an 80:20 training/evaluation split; this stage took approximately 36 hours, and the resulting accuracy is reported in Table~\ref{tab:vqa_acc}.

\subsubsection{Baselines}
Based on the results in Figure~\ref{fig:qualitative} and Table~\ref{tab:quantitative}, this paper employs Stable Diffusion v1.5~\cite{rombach2022high} as the baseline model due to its balanced performance in both image quality and text-image alignment compared to other versions. 

\subsubsection{Evaluation metrics}
To evaluate the quality of the generated images, we adopt five metrics that encompass both image quality and text-image alignment. For image quality assessment, we employ Fréchet Inception Distance (FID)~\cite{heusel2017gans} and LPIPS~\cite{zhang2018unreasonable}. For text-image alignment, we leverage CLIPScore~\cite{hessel2021clipscore} and BERTScore~\cite{zhang2019bertscore}. While these four metrics are widely established as standard benchmarks for evaluating diffusion-based sketch generation models~\cite{wang2023sketchknitter,hu2024scaleadaptive,vinker2022clipasso}, they are not specifically tailored to sketch abstraction.

We therefore additionally incorporate TIFAScore~\cite{hu2023tifa} and complement the quantitative evaluation with a user study.  This inclusion allows us to explicitly measure fine-grained prompt fidelity, providing a more precise assessment of semantic alignment than standard metrics alone. Finally, recognizing that these automated metrics may not fully capture the unique characteristics of sketches, we complemented our quantitative evaluation with a user study.

\input{tbl/quantitative}

\subsection{Preliminary experiments}
Initially, we evaluated mPLUG-large on the SketchDUO test sets and obtained an accuracy of 61.3\%, as reported in Table~\ref{tab:vqa_acc}. Since this performance was insufficient for reliable reward modeling, we fine-tuned mPLUG-large on SketchDUO. The fine-tuned model achieved a substantially improved accuracy of 89.3\% in the best setting. Furthermore, the model trained on both positive and negative samples consistently outperformed the version trained on positive samples alone, indicating that negative samples provide useful contrastive supervision for distinguishing desired sketch properties from undesired stylistic attributes.

Figure~\ref{fig:abl_tifascore} demonstrates that TIFAScore is more suitable than BERTScore for evaluating prompt fidelity, as it better captures the alignment between the text prompt and fine-grained elements of the generated image. While BERTScore focuses on overall semantic similarity, TIFAScore evaluates element-level fidelity, ensuring a more accurate assessment of how well the generated images meet the prompt's specific requirements.

\subsection{Quantitative results}
\input{fig/abl_tifascore/item}
\begin{figure}[!t]
\centering
\includegraphics[width=\columnwidth]{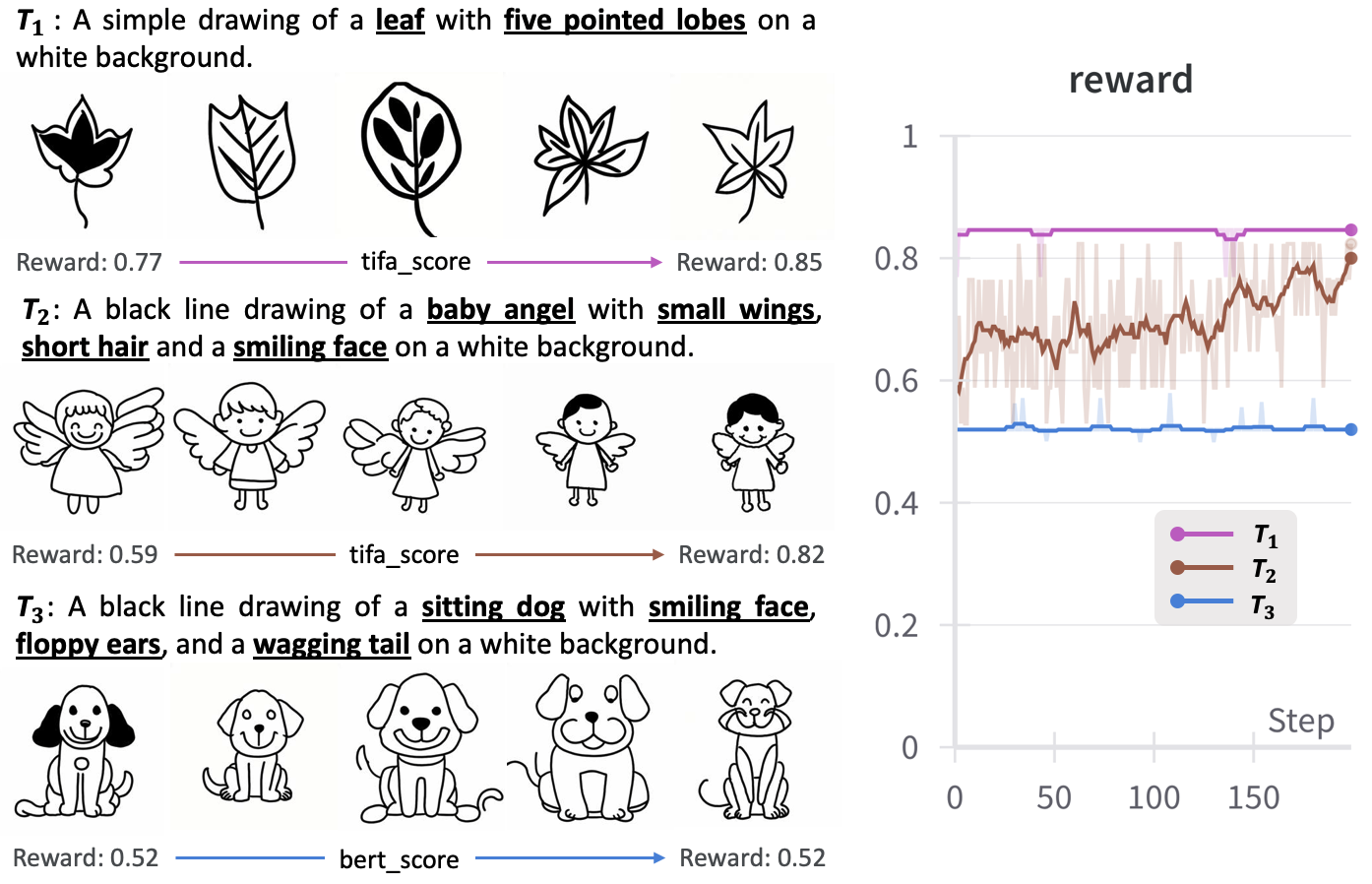}
\caption{Comparison of the progression of the DDPO algorithm with the new reward function. 
The left side illustrates the changes in generated images as the DDPO algorithm progresses for two sample prompts using the TIFAScore reward and one sample prompt using the BERTScore reward, while the right graph visualizes the reward progression for the respective prompts over the training steps.}
\label{fig:ddpo_result}
\end{figure}
\input{fig/qualitative/item}

Based on Table~\ref{tab:quantitative}, Stable Diffusion v1.5 demonstrated superior baseline performance compared to v2.1. While both models showed improvements with U-Net fine-tuning, v1.5 achieved greater enhancements in reducing FID and increasing TIFAScore. With additional VAE fine-tuning, v1.5 recorded the lowest FID of $143.68$ and the highest TIFAScore of $0.68$, delivering the best overall results. 

In contrast, VAE fine-tuning had minimal impact on v2.1's performance. Therefore, Stable Diffusion v1.5 with U-Net and VAE fine-tuning, which offers the best performance in text-image alignment and image quality, was selected for use with the DDPO algorithm.

Figure~\ref{fig:ddpo_result} illustrates the training process of the DDPO algorithm with the proposed reward function. For the text prompts \(T_1\) and \(T_2\), the generated images progressively aligned better with the prompts as training progressed. The reward for \(T_1\) increased from $0.77$ to $0.85$, while the reward for \(T_2\) improved from $0.59$ to $0.82$.
For \(T_1\), the reward initially increased rapidly and then stabilized, with the generated images progressively reflecting the finer details of the prompt. In contrast, \(T_2\) showed a steady improvement in the reward function throughout training, and the corresponding generated images consistently aligned more closely with the prompt.

Meanwhile, for prompt \(T_3\), the reward remained nearly unchanged throughout the training process, indicating limitations in achieving full prompt fidelity in the generated images.
The right graph of Figure~\ref{fig:ddpo_result} visualizes the reward progression over the training steps, demonstrating that the proposed reward function effectively enhances text-image alignment and stabilizes the learning process. The DDPO algorithm consistently generates images with higher prompt fidelity as training progresses, validating the effectiveness of the proposed reward function.

\subsection{Qualitative results}
Figure~\ref{fig:qualitative} compares image quality across Stable Diffusion variants and our framework, \emph{StableSketcher}. Stable Diffusion v1.5 and v2.1 show characteristic failures, as v1.5 often produces overly detailed outputs or drifts from the text prompt, whereas v2.1 tends to generate more abstract yet inconsistent results. Fine-tuning the U-Net improves simplicity and prompt alignment; tuning both U-Net and VAE further boosts fidelity but remains unstable for the “white background” and fully accurate instance generation.

\input{fig/userstudyscreen/item}
In contrast, the proposed StableSketcher applies DDPO, an RL-based policy optimization algorithm, to overcome the limitations of baseline models and achieve the best results. The images generated by StableSketcher resemble human-drawn sketches and faithfully reflect the detailed conditions of the text prompts. Additionally, the results from StableSketcher were the most similar to the ground truth images, demonstrating the effectiveness of the proposed framework.

\subsection{User study}

\input{tbl/user_study}

Figure~\ref{fig:userstudyscreen} shows the Google Forms-based interface used in our user study. The evaluation comprised three sections: Sketch Characteristics, Prompt Fidelity, and Human-Likeness. In the Sketch Characteristics section, participants assessed whether each image conformed to our definition of a sketch, with emphasis on simplicity, abstraction, black line drawing on a white background, and the absence of color, shading, and texture. In the Prompt Fidelity section, participants evaluated how well each image reflected the elements explicitly specified in the prompt, including object identity, attributes, and background conditions. In the Human-Likeness section, participants judged the extent to which each image resembled a human-drawn sketch rather than an overly synthetic or mechanically rendered output.
Each section consisted of 10 questions. To ensure full coverage of SketchDUO, we randomly partitioned its 30 classes into three disjoint sets of 10 classes and assigned one set to each section. For each class, the image used in the user study was randomly sampled from the corresponding class instances. In this way, all 30 classes were included exactly once in the overall study, while the evaluated examples were randomly selected.

We conducted a ranking-based user study with 46 participants. For each question, participants compared five model outputs and assigned ranks from 1 to 5, where 1 indicated the most preferred result and 5 indicated the least preferred result under the corresponding criterion. Table~\ref{tab:user_study} reports the mean rank for each method, where lower values indicate stronger user preference. Columns (a)--(d) correspond to Stable Diffusion v1.5, Stable Diffusion v2.1, Stable Diffusion v1.5 with U-Net fine-tuning, and Stable Diffusion v1.5 with both U-Net and VAE fine-tuning, respectively.
Our method achieved the lowest mean rank across all criteria, obtaining 1.9 for Sketch Characteristics, 1.7 for Prompt Fidelity, and 1.7 for Human-Likeness, with an overall mean rank of 1.7. Among the baselines, the strongest performance was obtained by the model in column (d), which achieved mean ranks of 2.3, 2.2, and 2.2 on the three criteria, respectively, and 2.2 overall. Compared with this baseline, our method improved the mean rank by 0.4 for Sketch Characteristics and by 0.5 for both Prompt Fidelity and Human-Likeness, resulting in an overall improvement of 0.5. These results indicate that participants consistently preferred our outputs over the baselines in terms of sketch style, prompt alignment, and perceived human-likeness.

\subsection{Ablation on VAE Loss Functions}
We evaluated different loss combinations for VAE fine-tuning through both reconstruction and generation tasks. For reconstruction, input sketches were encoded and decoded; for generation, the fine-tuned VAE was integrated into Stable Diffusion to produce sketches from text prompts (e.g., ``A black line drawing of a teddy bear with a friendly smile on a white background.'').

\begin{figure}[!t]
    \centering
    \includegraphics[width=1\linewidth]{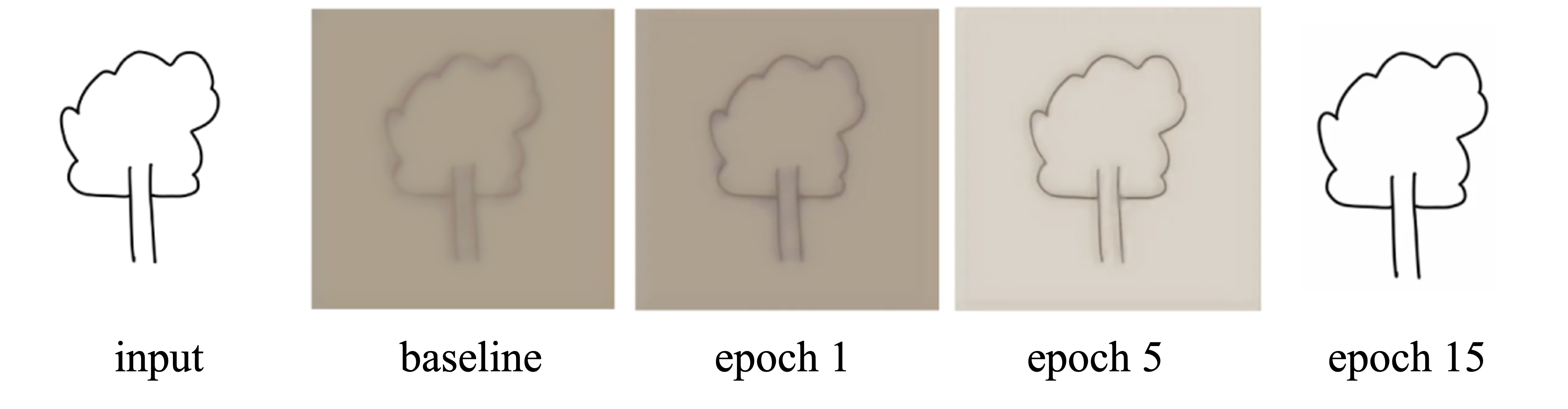}
    \caption{Reconstruction quality improvement over 15 epochs with our $L_{VAE} = L_{recon} + 10^{-1} \cdot L_{LPIPS}$, the combination of MSE loss and LPIPS loss.}
    \label{fig:mse_lpips}
\end{figure}

\begin{figure}[!t]
    \centering
    \includegraphics[width=1\linewidth]{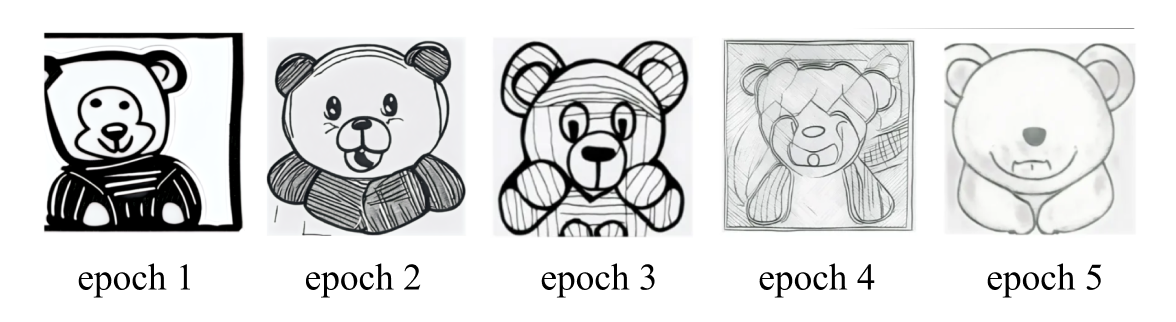}
    \caption{Image generation results across epochs 1 to 15 with MSE and KL combination loss.}
    \label{fig:white}
\end{figure}

\begin{figure*}[!t]
\centering
\includegraphics[width=0.8\linewidth]{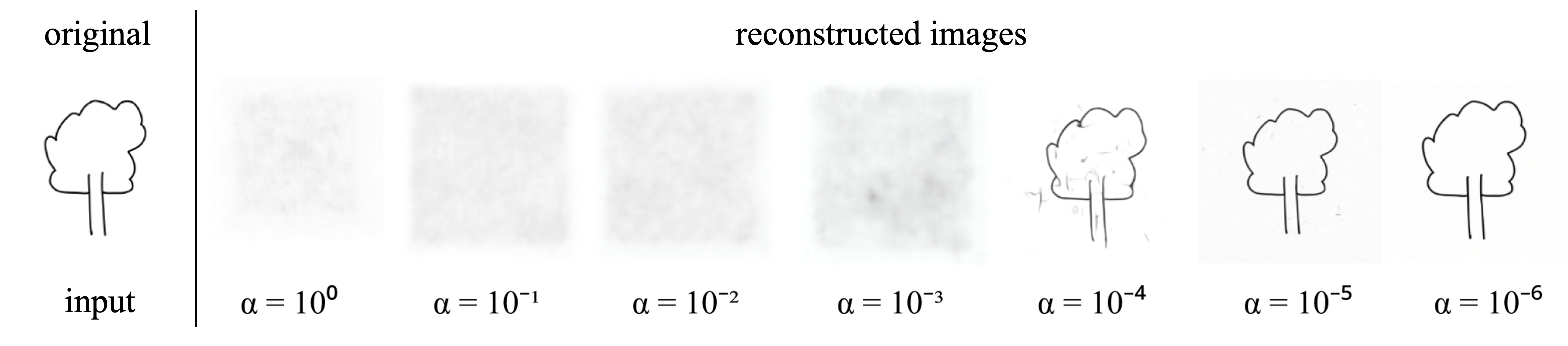}
\caption{Effect of KL weight on VAE reconstruction quality. In the reconstructed images, $\alpha$ denotes the coefficient in the VAE loss, $\mathrm{L_{recon}} + \alpha\cdot\mathrm{L_{KL}}$.}
\label{fig:kl_ablation}
\end{figure*}

\begin{figure*}[!t]
\centering
\includegraphics[width=0.8\linewidth]{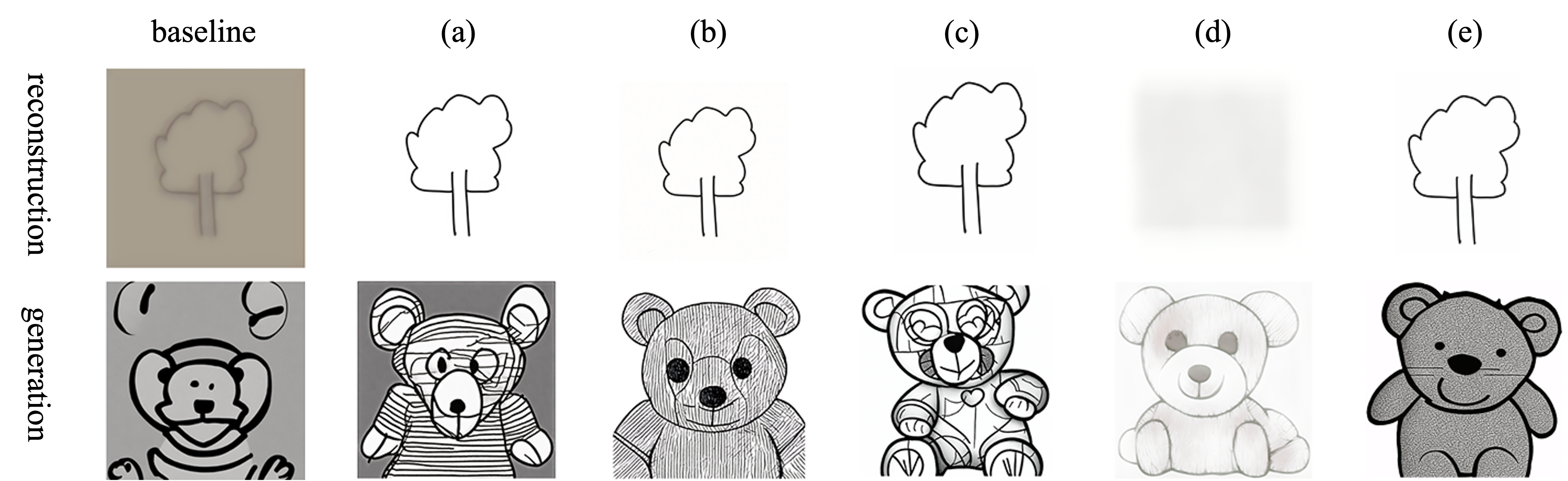}
\caption{Qualitative comparison of reconstruction and generation after 15 epochs under different loss compositions. (a)–(e) correspond to $L_{recon}$, $L_{recon}+10^{-6}\cdot L_{KL}$, $L_{recon}+10^{-1}\cdot L_{LPIPS}$, $L_{recon}+10^{-1}\cdot L_{KL}+10^{-1}\cdot L_{LPIPS}$, and $L_{L1}+10^{-1}\cdot L_{LPIPS}$, respectively.}
\label{fig:losscombi}
\end{figure*}

\paragraph{Effect of MSE and LPIPS}
When MSE was combined with LPIPS, reconstruction quality improved steadily over 15 epochs, as illustrated in Figure~\ref{fig:mse_lpips}. Beyond visual gains, this combination also produced consistent reductions in both pixel-wise and perceptual errors: after 15 epochs, the MSE decreased from $0.0008$ to $0.00017$ and the LPIPS from $0.0028$ to $0.0005$. These results show that MSE preserves low-level accuracy, while LPIPS enforces perceptual consistency in contours and line structures, leading to more stable training and improved sketch reconstruction.

\paragraph{Effect of KL Divergence}
When training the VAE with a combination of MSE and KL loss, as in the original formulation, the generated outputs gradually collapsed into almost entirely white backgrounds, as shown in Figure~\ref{fig:white}. This occurs because a large KL weight over-regularizes the latent space, forcing the encoder to map inputs too closely to a standard normal distribution and thereby discarding fine-grained sketch details. As a result, the model fails to reconstruct the black line structures of the sketches, which is consistent with the findings reported in Appendix G of Stable Diffusion~\cite{rombach2022high}. 

We also examined the effect of varying the KL weight across several orders of magnitude, from $10^{0}$ down to $10^{-6}$. When the weight was set to $10^{-6}$, the KL divergence loss exploded while the reconstruction loss remained low, indicating unstable training. In contrast, setting the weight to $10^{0}$ caused the KL loss to nearly vanish but led to a collapse in reconstruction quality. These results, illustrated in Figure~\ref{fig:kl_ablation}, confirm that improper weighting of the KL term severely degrades the VAE’s ability to preserve sketch information.

\paragraph{Other Loss Variants}
We additionally considered binary cross-entropy (BCE) loss; however, its constraint that outputs lie in $[0,1]$ is incompatible with the LPIPS objective and our VAE decoder configuration. Optimizing with LPIPS alone yielded thick yet consistent contours, whereas coupling L1 with LPIPS delivered smaller perceptual gains than the MSE--LPIPS pairing and left LPIPS effectively unchanged ($\approx 1\times10^{-3}$) after 15 epochs. Taken together, the MSE--LPIPS combination offered the most favorable trade-off between pixel-level fidelity and perceptual sketch quality, as summarized in Figure~\ref{fig:losscombi}.

%% file: tbl/quantitative.tex
\newcommand{\mstd}[2]{#1{\tiny$\pm$#2}}
\newcommand{\bmstd}[2]{\textbf{#1}{\tiny$\pm$#2}}

\begin{table*}[t]
\centering
\caption{Quantitative evaluation of generated images using FID, LPIPS, CLIPScore, BERTScore, and TIFAScore metrics for different configurations of Stable Diffusion models. The ``+ Fine-tuning'' rows indicate that fine-tuning was applied to the corresponding base model, while ``+ VAE fine-tuning'' rows represent the additional application of VAE fine-tuning on top of the fine-tuned model.}
\label{tab:quantitative}
\renewcommand{\arraystretch}{1.3}
\setlength{\tabcolsep}{8pt}
\begin{tabular}{lccccc}
\toprule
\textbf{Method} & \textbf{FID $\downarrow$} & \textbf{LPIPS $\downarrow$} & \textbf{CLIPScore $\uparrow$} & \textbf{BERTScore $\uparrow$} & \textbf{TIFAScore $\uparrow$} \\
\midrule
Stable Diffusion v1.5 & \mstd{207.59}{22.29} & \mstd{0.69}{0.09} & \mstd{34.00}{2.59} & \bmstd{0.89}{0.03} & \mstd{0.59}{0.15} \\
\quad + U-Net fine-tuning & \mstd{161.94}{20.33} & \mstd{0.40}{0.09} & \bmstd{36.05}{2.59} & \bmstd{0.89}{0.03} & \bmstd{0.68}{0.13} \\
\qquad + VAE fine-tuning & \bmstd{143.68}{16.58} & \bmstd{0.37}{0.08} & \mstd{35.48}{2.50} & \mstd{0.88}{0.03} & \bmstd{0.68}{0.14} \\
\midrule
Stable Diffusion v2.1 & \mstd{230.78}{22.65} & \mstd{0.72}{0.07} & \mstd{31.13}{3.42} & \mstd{0.88}{0.03} & \mstd{0.53}{0.15} \\
\quad + U-Net fine-tuning & \mstd{144.46}{25.68} & \mstd{0.41}{0.07} & \mstd{34.79}{2.71} & \mstd{0.88}{0.03} & \mstd{0.67}{0.13} \\
\qquad + VAE fine-tuning & \mstd{172.35}{14.48} & \mstd{0.50}{0.08} & \mstd{34.11}{2.84} & \mstd{0.88}{0.03} & \mstd{0.65}{0.13} \\
\bottomrule
\end{tabular}
\end{table*}

%% file: fig/abl_tifascore/item.tex
\begin{figure}[t!]
\centering
\includegraphics[width=0.99\linewidth]{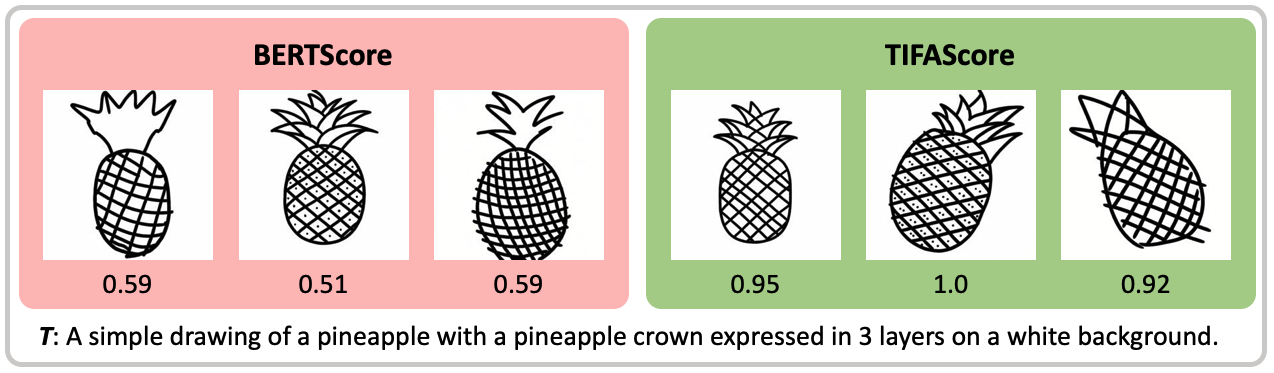}
\caption{BERTScore and TIFAScore evaluations for generated images based on the text prompt describing a "simple drawing of a pineapple with a crown expressed in 3 layers on a white background.
}
\label{fig:abl_tifascore}
\end{figure}

%% file: fig/qualitative/item.tex
\begin{figure*}[!t]
\centering
\includegraphics[width=0.99\linewidth]{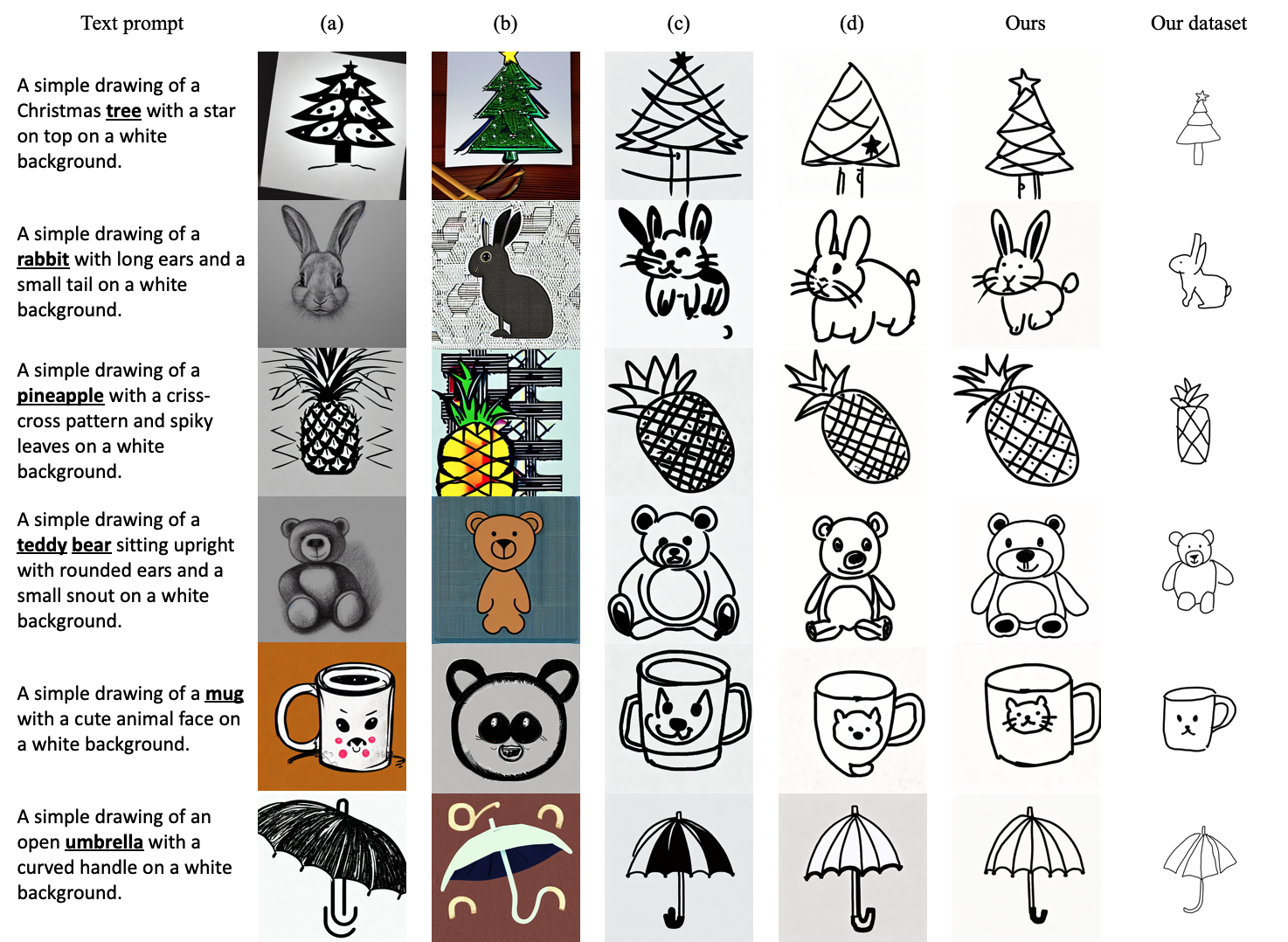}
% \vspace{-0.2cm}
\caption{
Qualitative comparison of images generated by different models based on the input text prompts. (a) Images generated by Stable Diffusion v1.5, baseline model. (b) Images generated by Stable Diffusion v2.1. (c) Outputs from fine-tuning only the UNet component of Stable Diffusion v1.5. (d) Outputs from fine-tuning both the UNet and VAE components of Stable Diffusion v1.5. ``Ours" represents the results from our proposed framework, StableSketcher. ``Our dataset" displays the ground truth images corresponding to the text prompts. Each example illustrates a representative class from six categories.}
% \vspace{-0.4cm}
\label{fig:qualitative}
\end{figure*} 

%% file: fig/userstudyscreen/item.tex
\begin{figure*}[!t]
\centering
\includegraphics[width=0.99\linewidth]{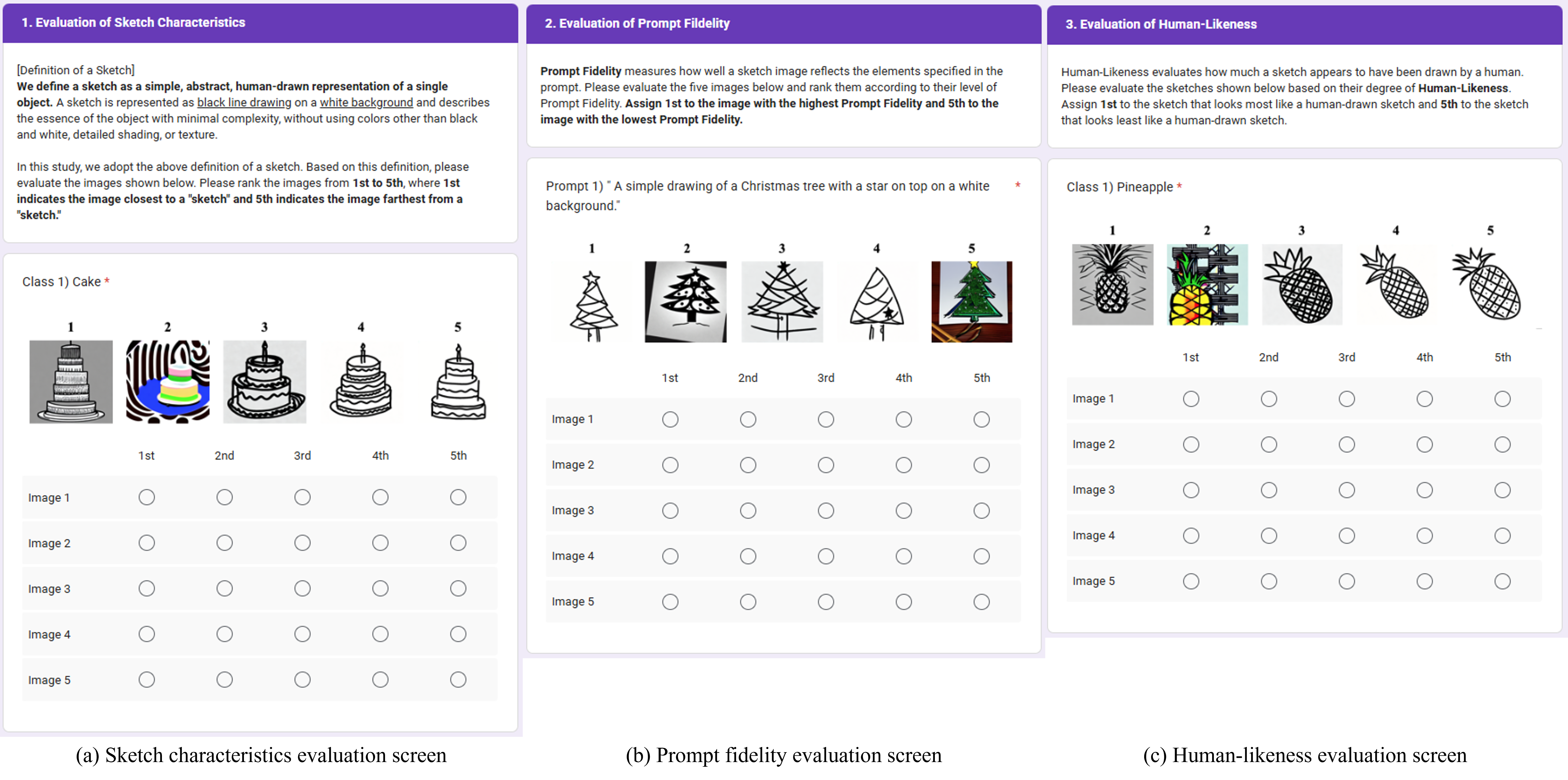}
\caption{
User study interface used for subjective evaluation.
The user study was conducted using a Google Forms interface and consisted of three evaluation criteria: (a) sketch characteristics, (b) prompt fidelity, and (c) human-likeness. For each criterion, participants were presented with five generated images and asked to rank them from 1st to 5th according to the corresponding evaluation guideline.}
\label{fig:userstudyscreen}
\end{figure*} 

%% file: tbl/user_study.tex
\begin{table}[!t]
\caption{User study results for each model corresponding to the visual samples in Figure~\ref{fig:qualitative}. Here, (a) denotes images generated by Stable Diffusion v1.5, (b) denotes images generated by Stable Diffusion v2.1, (c) denotes outputs from fine-tuning only the UNet component of Stable Diffusion v1.5, and (d) denotes outputs from fine-tuning both the UNet and VAE components of Stable Diffusion v1.5.}
\label{tab:user_study}
\centering
\begin{tabular}{lccccc}
\toprule
\textbf{Criterion} (Mean Rank ↓) & \textbf{(a)} & \textbf{(b)} & \textbf{(c)} & \textbf{(d)} & \textbf{Ours} \\
\midrule
Sketch Characteristics & 3.8 & 4.1 & 2.7 & 2.3 & \textbf{1.9} \\
Prompt Fidelity        & 4.3 & 4.1 & 2.4 & 2.2 & \textbf{1.7} \\
Human-Drawn            & 4.0 & 4.4 & 2.5 & 2.2 & \textbf{1.7} \\
\midrule
Total Average Rank     & 4.0 & 4.2 & 2.5 & 2.2 & \textbf{1.7} \\
\bottomrule
\end{tabular}
\end{table}

%% file: tex/6_conclusion.tex
\section{Discussion}
The quantitative results demonstrate the effectiveness of our framework. \textit{StableSketcher} achieved the lowest FID (143.68) and the highest TIFAScore (0.68) across all configurations, outperforming the Stable Diffusion v1.5 and v2.1 baselines. These gains indicate that the proposed reconstruction--perceptual hybrid loss and VQA-based reinforcement learning improve alignment between generated sketches and fine-grained textual prompts.

BERTScore showed negligible differences between our model and the baselines. This suggests that, although BERTScore captures the overall semantic impression of an image, it is less effective at evaluating whether specific element-level conditions in the prompt are faithfully reflected in the output. This observation further supports the need for our VQA-based reward function in improving prompt fidelity.

Another notable finding is that Stable Diffusion v1.5 outperformed v2.1 in sketch generation. While v2.1 is optimized for high-quality photorealistic synthesis through more aggressively filtered training data~\cite{sd2_release}, this bias appears to hinder abstraction-oriented generation. By contrast, v1.5 was trained on a broader and less aggressively filtered distribution, including LAION-2B(en) and LAION-Aesthetics v2.5+ subsets~\cite{sd15_modelcard}, which may make it more adaptable to non-photorealistic domains. This distributional difference likely explains the stronger performance of the v1.5-based \textit{StableSketcher}.

Despite these advances, our study still has several limitations. \textit{SketchDUO} remains relatively small, containing only 30 categories and 4.7K sketches, and it also exhibits a Western-centric bias, which may limit generalization across broader domains and more diverse cultural contexts. In addition, our evaluation is conducted within the controlled style space defined by \textit{SketchDUO}, and robustness to broader external human-drawn sketch distributions remains an important direction for future work. Our quantitative evaluation also relies on standard image-generation metrics, including FID, CLIPScore, and BERTScore. Although these metrics are widely used in recent diffusion-based sketch generation studies and support direct comparison with prior work, they are not specifically designed to capture sketch abstraction. To mitigate this limitation, we complemented the quantitative evaluation with a user study and are currently exploring dedicated evaluation metrics in follow-up work. Finally, the annotation pipeline would benefit from more scalable semi-automated strategies, which may improve annotation efficiency while enabling broader category coverage and more diverse sketch content. Although our current setup is not fully dependent on closed-source systems, GPT-4o was used only for initial caption drafting, whereas QA pairs were constructed using the open-source TIFA pipeline with LLaMA 2 and UnifiedQA. Reducing reliance on closed-source captioning models and developing stronger open-source alternatives remain important directions for future work.

\section{Conclusion}
In this work, we proposed \textit{StableSketcher}, a framework for generating human-drawn, pixel-based sketches with Stable Diffusion, supported by \textit{SketchDUO}, a novel dataset containing triplets of sketch images, fine-grained captions, and QA pairs. By incorporating positive examples of desired abstraction and negative examples of common errors, such as over-shading or photorealistic bias, our contrastive design helps disentangle semantic correctness from stylistic faithfulness.

Beyond its technical contributions, \textit{StableSketcher} may support interactive sketch-generation applications. For example, it could support co-drawing with an artificial intelligence assistant for older adults or be used in creative drawing activities for children through iterative sketch expansion. More broadly, our work suggests the potential of sketch-oriented generative models for applications in education, therapy, and lifelong learning.

To address the limitations identified in this study, we are currently undertaking follow-up research. We are constructing a larger and more diverse dataset encompassing 300 categories to alleviate the scale and cultural constraints of the current dataset. At the same time, we are developing novel evaluation metrics designed to better capture sketch abstraction and stylistic fidelity beyond general-purpose image metrics. These efforts aim to provide a stronger foundation for sketch-oriented text-to-image generation.

%% file: tex/7_acknowledgment.tex
\section*{Acknowledgment}
\small
This research was supported by the ‘Regional Innovation System \& Education (RISE)' through the Seoul RISE Center, funded by the Ministry of Education (MOE) and the Seoul Metropolitan Government (2025-RISE-01-007-05); and the Institute of Information \& Communications Technology Planning \& Evaluation (IITP) under the Artificial Intelligence Convergence Innovation Human Resources Development grant funded by the Korea government (MSIT) (IITP-2025-RS-2023-00254592). Additionally, this work has been carried out with extensive use of the NEURON computing resource supported by the Korea Institute of Science and Technology Information (KISTI).
This study involved human participants. All procedures were approved by the Institutional Review Board (IRB) of Dongguk University (Approval No. DUIRB-2025-05-08) and were conducted in accordance with institutional ethical guidelines.

%% file: tex/8_appedix.tex
\clearpage
\onecolumn
\section*{Appendix: Qualitative Evaluation Results}
% \label{app:qual}
% (Optional) If you want appendix-specific numbering like "Fig. A1, A2, ...", uncomment below:
% \renewcommand{\thefigure}{A\arabic{figure}}
% \setcounter{figure}{0}
\begin{figure*}[!h]
    \centering
    \includegraphics[width=0.76\linewidth, keepaspectratio]{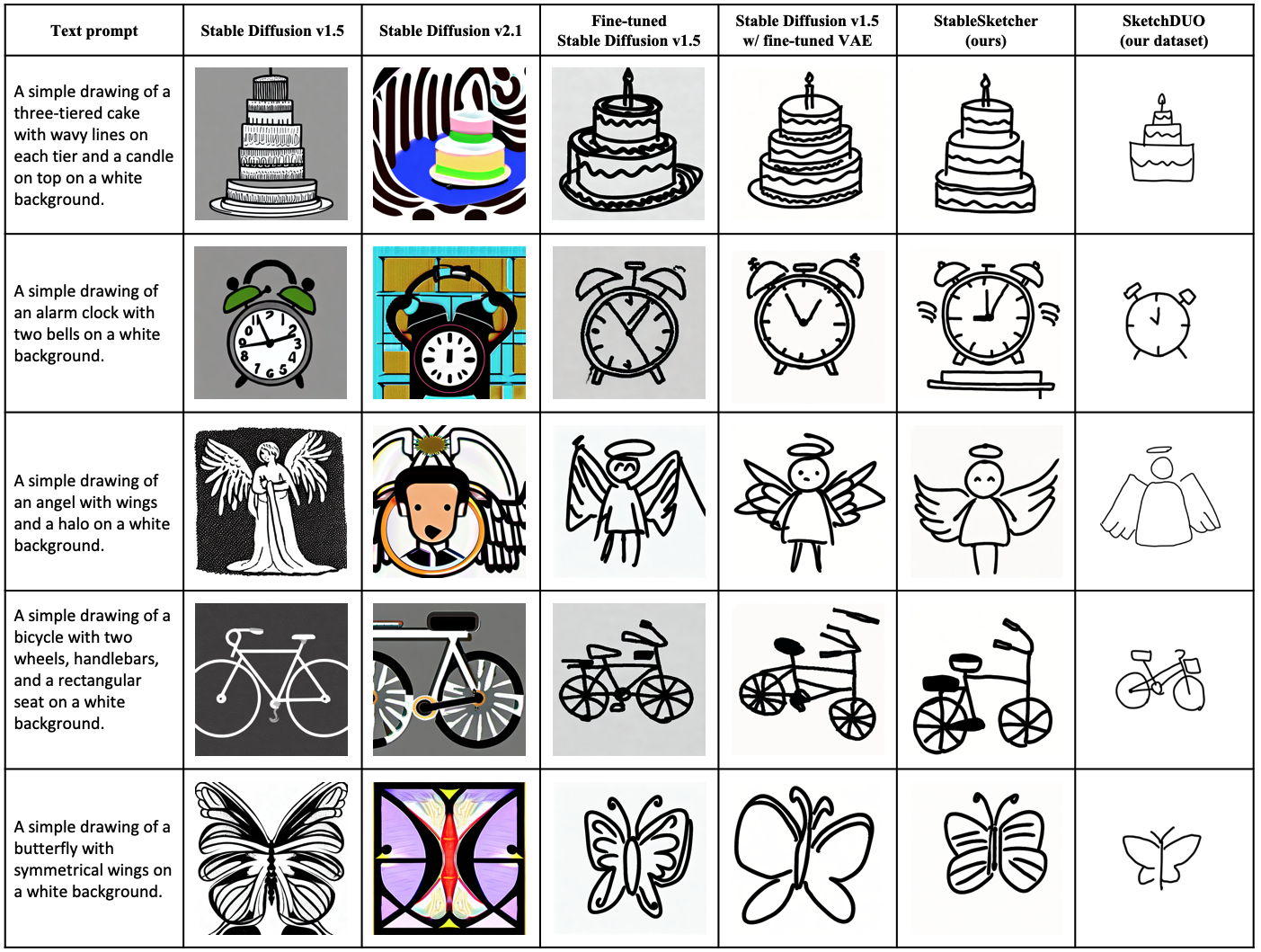}

    \medskip

    \includegraphics[width=0.76\linewidth, keepaspectratio]{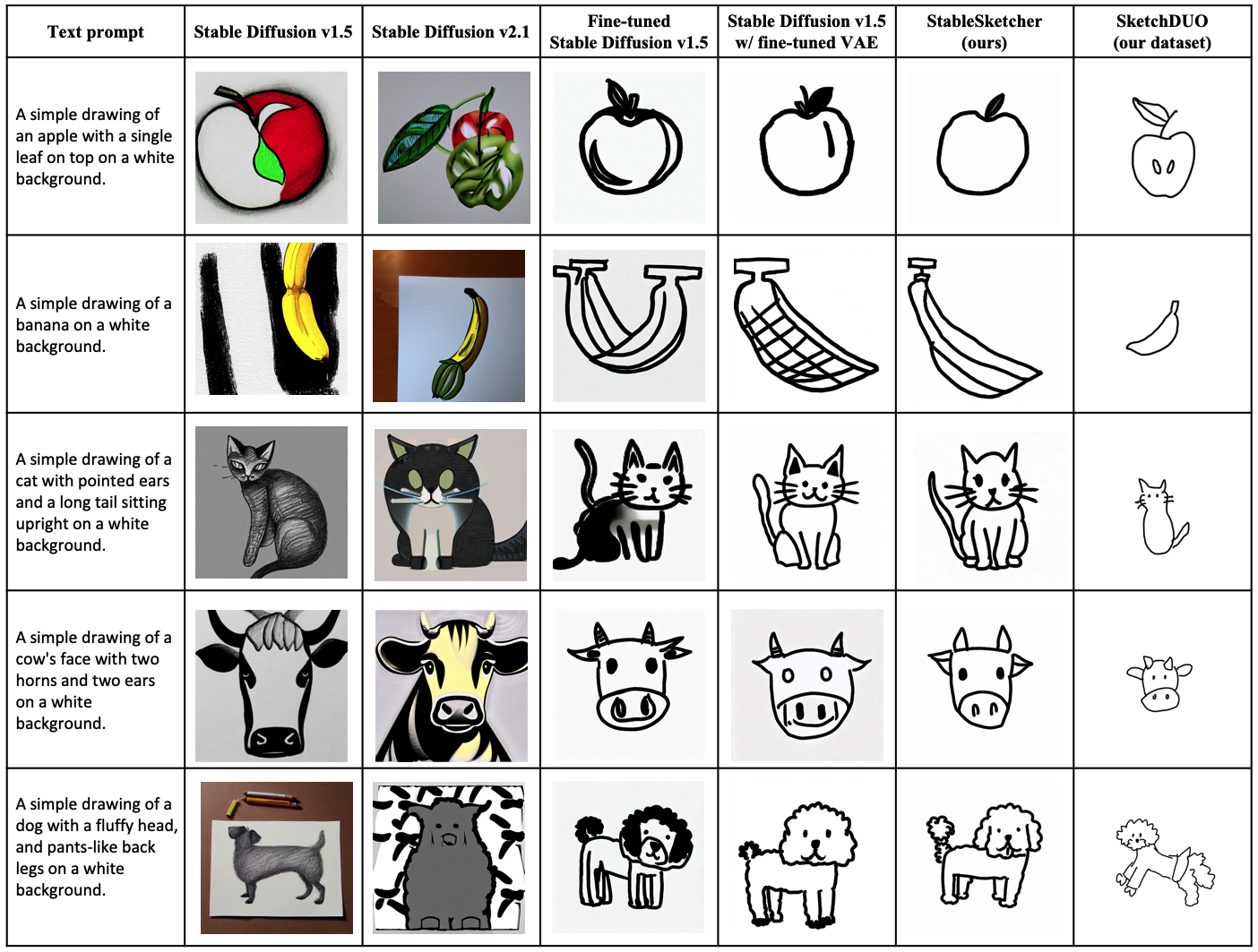}
    \caption{Qualitative comparison of images generated by different models based on the input text prompts. “Ours” denotes results from the proposed StableSketcher. “Our dataset” shows the ground-truth images corresponding to the prompts.}
    \label{fig:total_combined12}
\end{figure*}

\begin{figure*}[h]
    \centering
    \includegraphics[width=0.76\linewidth, keepaspectratio]{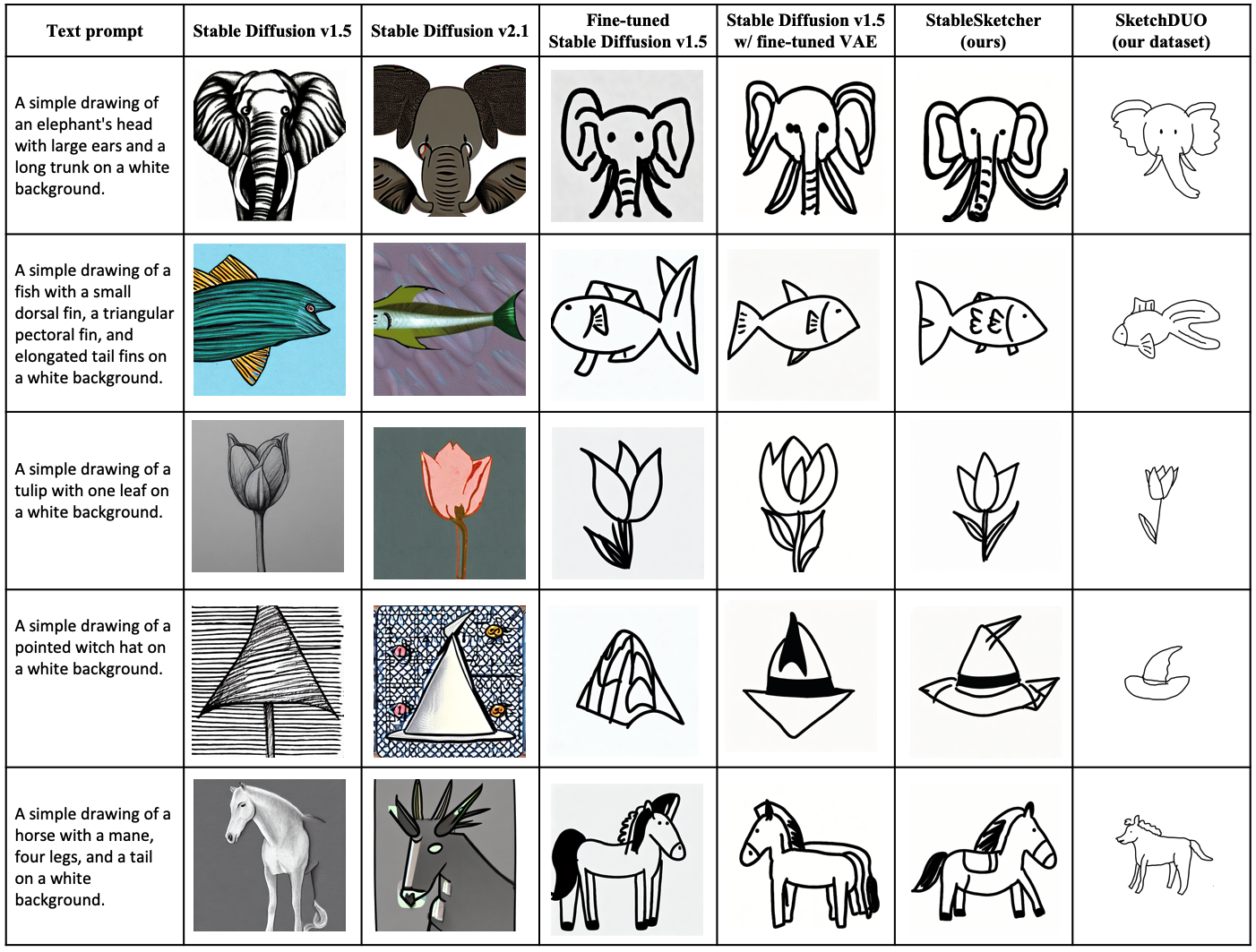}

    \medskip

    \includegraphics[width=0.76\linewidth, keepaspectratio]{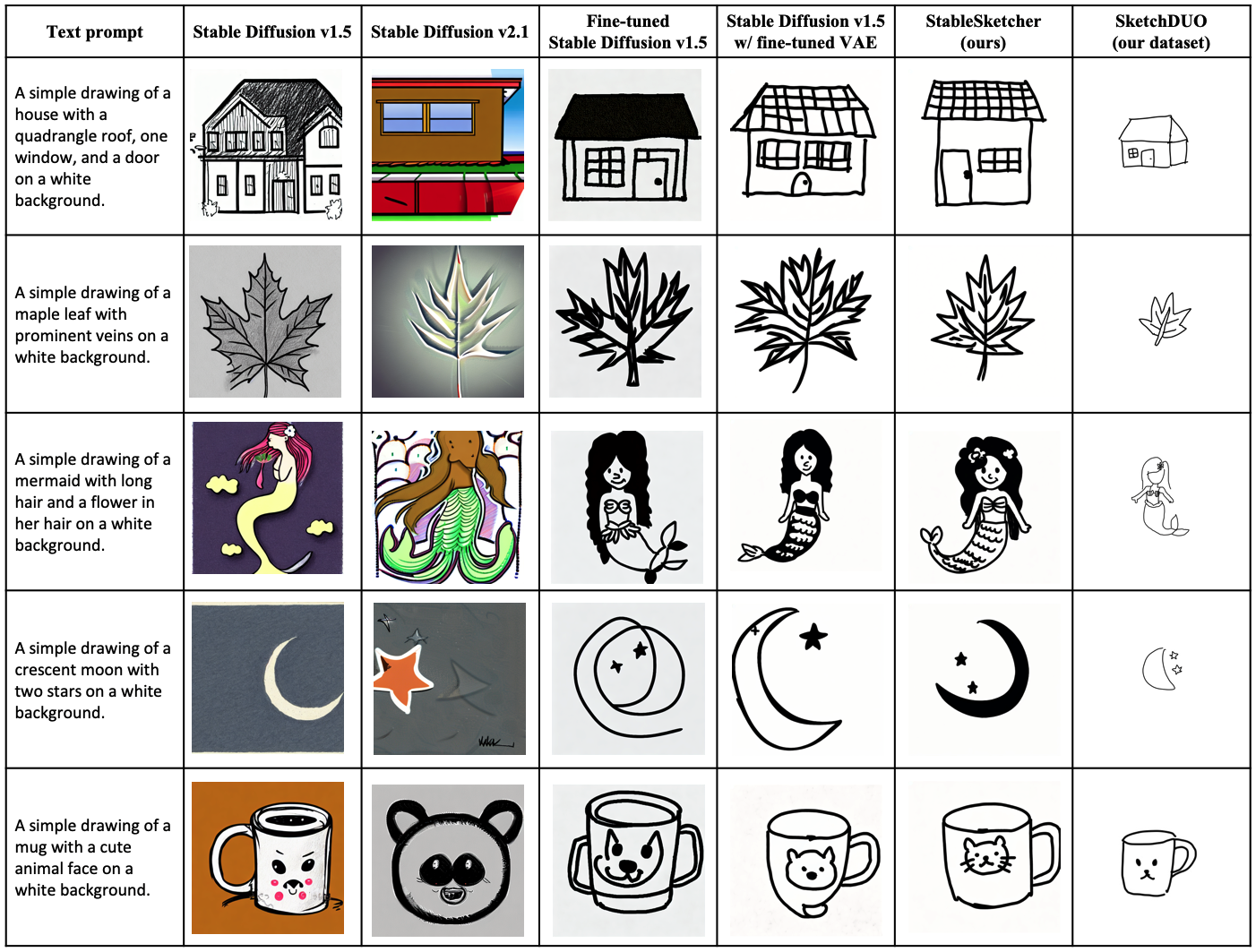}
    \caption{Qualitative comparison based on input text prompts (set 2). “Ours” denotes results from StableSketcher; “Our dataset” shows the corresponding ground-truth images.}
    \label{fig:total_combined34}
\end{figure*}

\begin{figure*}[!t]
    \centering
    \includegraphics[width=0.76\linewidth, keepaspectratio]{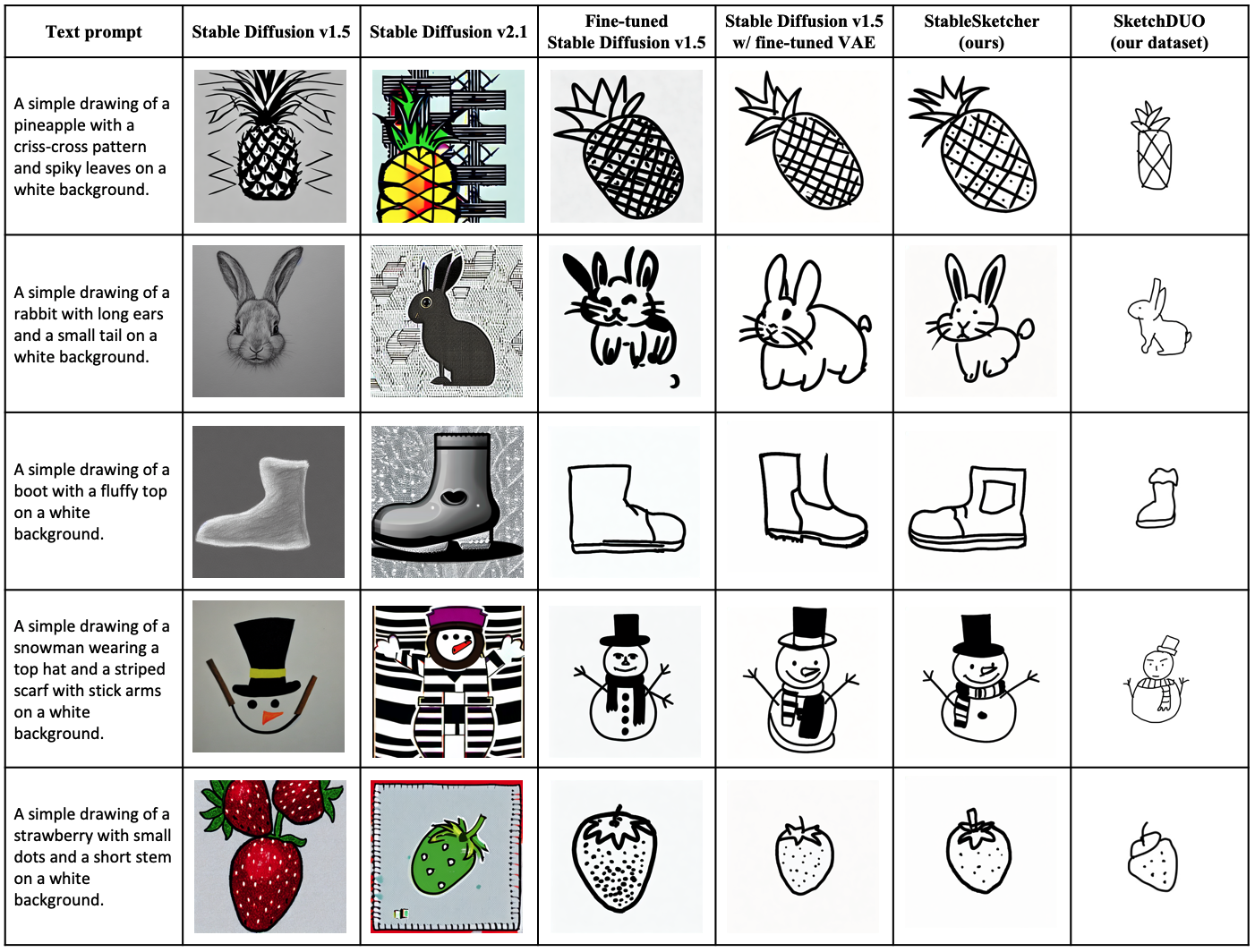}

    \medskip

    \includegraphics[width=0.76\linewidth, keepaspectratio]{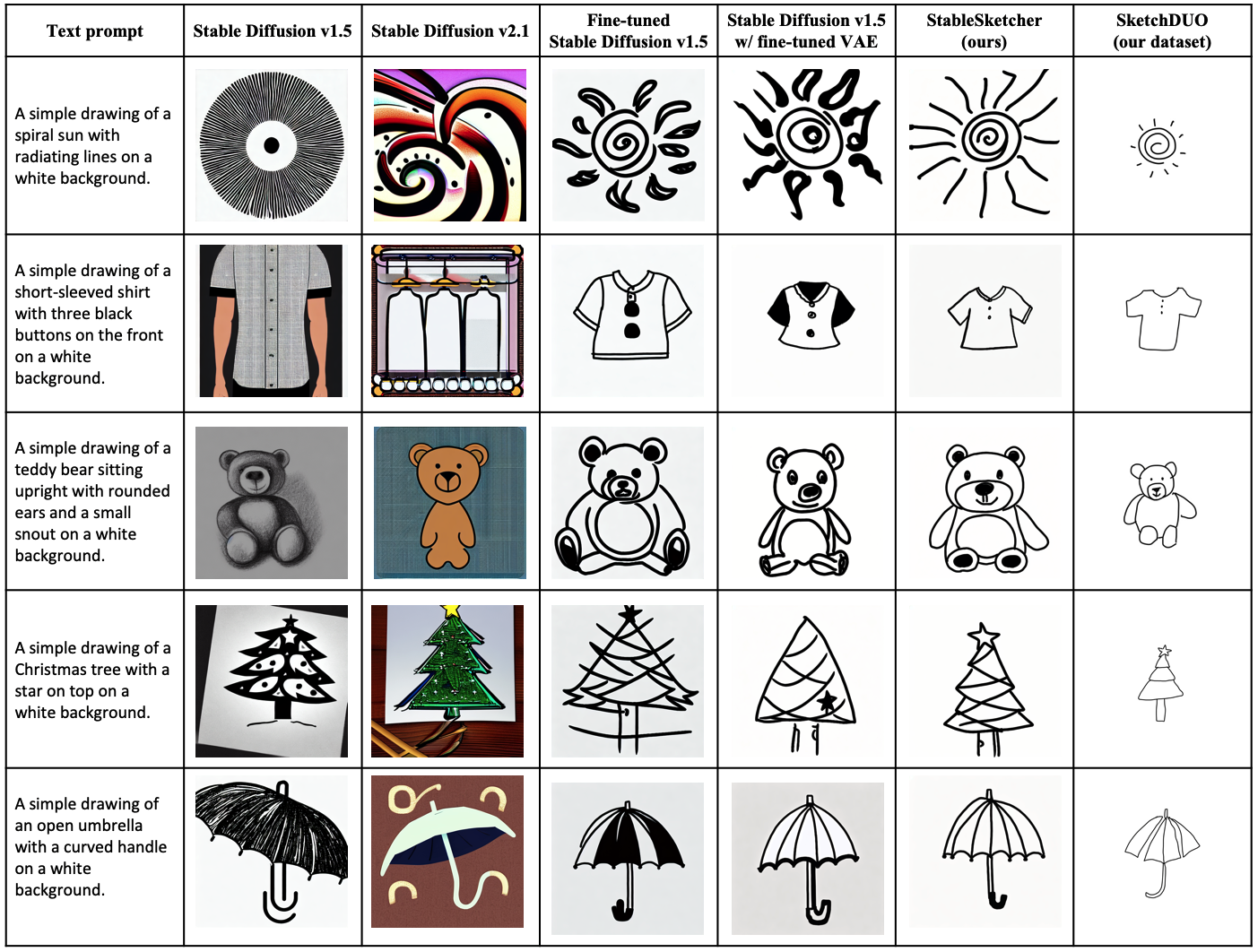}
    \caption{Qualitative comparison based on input text prompts (set 3). “Ours” denotes results from StableSketcher; “Our dataset” shows the corresponding ground-truth images.}
    \label{fig:total_combined56}
\end{figure*}

\clearpage
\twocolumn